\documentclass{article}

\PassOptionsToPackage{numbers, compress}{natbib}
\usepackage[preprint]{neurips_2024}

\usepackage[utf8]{inputenc} 
\usepackage[T1]{fontenc}    
\usepackage{hyperref}       
\usepackage{url}            
\usepackage{booktabs}       
\usepackage{amsfonts}       
\usepackage{nicefrac}       
\usepackage{microtype}      
\usepackage{xcolor}         

\usepackage{booktabs,multicol,multirow}

\usepackage{amsmath}
\usepackage{amsfonts}
\usepackage{mathrsfs}
\usepackage{dsfont}
\usepackage{stmaryrd}

\usepackage{microtype}
\usepackage{graphicx}
\usepackage{xcolor}
\usepackage{algorithm}
\usepackage[noend]{algorithmic}
\usepackage{caption,subcaption}
\usepackage{mathtools}
\usepackage{annotate-equations}
\usepackage{soul}
\usepackage{listings}

\newtheorem{definition}{Definition}[section]

\newcommand{\ie}{i.e.,\ }
\newcommand{\eg}{e.g.,\ }
\newcommand{\wrt}{w.r.t.\ }

\newcommand{\ontology}{\ensuremath{\mathcal{O}}}
\newcommand{\Co}{\ensuremath{\mathcal{C}_\ontology}}
\newcommand{\KB}{\ensuremath{\mathcal{KB}}}
\newcommand{\KG}{\ensuremath{\mathcal{KG}}}
\newcommand{\influeceRole}{\ensuremath{\mathsf{influence}}}
\newcommand{\signedRole}{\ensuremath{\mathsf{signedTo}}}
\newcommand{\artistConcept}{\ensuremath{\mathsf{Artist}}}
\newcommand{\labelConcept}{\ensuremath{\mathsf{Label}}}

\newcommand{\I}{\ensuremath{\mathcal{I}}}
\newcommand{\tbox}{TBox}
\newcommand{\abox}{ABox}
\newcommand{\concepts}{\ensuremath{\mathbf{C}}}
\newcommand{\roles}{\ensuremath{\mathbf{R}}}
\newcommand{\ALC}{\ensuremath{\mathcal{ALC}}}
\newcommand{\ALCI}{\ensuremath{\mathcal{ALCI}}}
\newcommand{\DeltaI}{\ensuremath{\Delta^{\I}}}

\newcommand{\X}{\ensuremath{\mathbf{X}}}
\newcommand{\cinput}[1]{\ensuremath{\mathsf{in}(#1)}}
\newcommand{\cscope}[1]{\ensuremath{\mathsf{sc}(#1)}}
\newcommand{\csupport}[1]{\ensuremath{\mathsf{supp}(#1)}}

\newcommand{\CI}{\ensuremath{C^{\I}}}
\newcommand{\AI}{\ensuremath{A^{\I}}}
\newcommand{\aI}{\ensuremath{a^{\I}}}
\newcommand{\BI}{\ensuremath{B^{\I}}}

\newcommand{\RI}{\ensuremath{R^{\I}}}

\newcommand{\arb}{\ensuremath{\langle a, b \rangle}}
\newcommand{\arbI}{\ensuremath{\langle a, b \rangle^{\I}}}

\newcommand{\braI}{\ensuremath{\langle b, a \rangle^{\I}}}
\newcommand{\domino}{\ensuremath{\langle \mathcal{A}, \mathcal{R}, \mathcal{B} \rangle}}
\newcommand{\parts}{\ensuremath{\mathsf{parts}}}

\newcommand{\canonicaldomino}{\ensuremath{\mathbb{D}_{\ontology}}}
\newcommand{\cbf}[1]{\llbracket #1 \rrbracket}
\newcommand{\Vars}{\mathbf{Vars}}
\newcommand{\CKB}{\ensuremath{\mathcal{C}_\KB}}

\newcommand{\nn}{\mathsf{nn}}

\newcommand{\dataset}{\mathcal{D}}

\newcommand{\yhat}{\mathbf{\Hat{y}}}

\newcommand{\bbR}{\mathbb{R}}
\newcommand{\argmax}{\mathop{\mathrm{argmax}}}

\title{To Neuro-Symbolic Classification and Beyond by Compiling Description Logic Ontologies to Probabilistic Circuits}

\author{%
    Nicolas Lazzari \\
    University of Pisa, University of Bologna \\
    Italy \\
    \texttt{nicolas.lazzari3@unibo.it}
    \And
    Valentina Presutti \\
    University of Bologna \\
    Italy \\
    \texttt{valentina.presutti@unibo.it}
    \And
    Antonio Vergari \\
    University of Edinburgh \\
    United Kingdom \\
    \texttt{avergari@ed.ac.uk}
}

\begin{document}
\maketitle

\begin{abstract}
    {\bf Background:} 
    Neuro-symbolic methods enhance the reliability of neural network classifiers through logical constraints, but they lack native support for ontologies.
    
    {\bf Objectives:}
    We aim to develop a neuro-symbolic method that reliably outputs predictions consistent with a Description Logic ontology that formalizes domain-specific knowledge.
    
    {\bf Methods:}
    We encode a Description Logic ontology as a circuit, a feed-forward differentiable computational graph that supports tractable execution of queries and transformations. We show that the circuit can be used to (i) generate synthetic datasets that capture the semantics of the ontology; (ii) efficiently perform deductive reasoning on a GPU; (iii) implement neuro-symbolic models whose predictions are approximately or provably consistent with the knowledge defined in the ontology.

    {\bf Results:}
    We show that the synthetic dataset generated using the circuit qualitatively captures the semantics of the ontology while being challenging for Machine Learning classifiers, including neural networks. Moreover, we show that compiling the ontology into a circuit is a promising approach for scalable deductive reasoning, with runtimes up to three orders of magnitude faster than available reasoners. Finally, we show that our neuro-symbolic classifiers reliably produce consistent predictions when compared to neural network baselines, maintaining competitive performances or even outperforming them.
    
    {\bf Conclusions:}
    By compiling Description Logic ontologies into circuits, we obtain a tighter integration between the Deep Learning and Knowledge Representation fields. We show that a single circuit representation can be used to tackle different challenging tasks closely related to real-world applications.
\end{abstract}

\section{Introduction}
Deep Learning methods achieved unprecedented performances in complex tasks such as understanding natural language \citep{sutskever2014seqtoseq,vaswani2017attention,brown2020gpt3,touvron2023llama} or images \citep{krizhevsky20212imagenet,he2016resnet,dosovitskiy2021vit,simeoni2025dino} by scaling models to billions of parameters and training them on large quantities of data. 

Their reliability, defined as the ability of the model to output predictions that are consistent with background knowledge, is a critical requirement in real-world applications but is often overlooked since it is difficult to propose reliable methods by design or to assess them once they are applied in the wild.

Suppose that we want to design a model for automated information extraction in a domain-specific setting. For instance, we are interested in automatically detecting the creative influence between different artists from a biographical document containing the sentence
\begin{equation}
    \text{``Fugazi, also Dischord Records, were a major influence to The Smiths.''}
    \label{ex:sentence}
\end{equation}
where \textit{Fugazi}\footnote{\url{https://en.wikipedia.org/wiki/Fugazi}} and \textit{The Smiths}\footnote{\url{https://en.wikipedia.org/wiki/The_Smiths}} are artists and \textit{Dischord Records}\footnote{\url{https://en.wikipedia.org/wiki/Dischord_Records}} is a recording label, as defined by some domain-specific background knowledge at hand. If we define artistic influence as a relation that holds only between artists, a reliable model \textit{never} predicts that \textit{Dischord Records} influenced \textit{The Smiths}, independently of the way the information is presented. One solution is to train models on large datasets that are consistent with background knowledge but exhibit a large variance with respect to their content. This approach, however, is difficult to pursue in domain-specific settings that are typically characterized by data scarcity \citep{hedderich2021lowreslanguage,zhang2024lowresvision}. 

A different approach is to integrate this knowledge directly into the model. This presents two main challenges: effectively representing the background knowledge and integrating it into complex neural networks. On the one hand, a large number of \textit{Knowledge Representation} (KR) methods have been proposed \citep{vanharmelen2008kr} to computationally encode knowledge into Knowledge Bases, generally relying on symbolic methods such as logic languages. Ontologies defined using Description Logic (DL), in particular, are one of the most widespread KR methods \citep{baader2003dlhandbook,vanharmelen2008kr,staab2004ontologieshandbook,brachman2004kr} due to their appealing tradeoff between computational complexity and expressivity.

On the other hand, the Neuro-Symbolic (NeSy) field investigates how to integrate methods developed in the KR field with neural networks \citep{manhaeve2021nesy,garcez2023nesyai} in the form of domain-specific constraints \citep{giunchiglia2022logicconstraints,morettin2021logic}. These approaches are generally characterized by a \textit{tension} between expressivity, tractability, and guaranteed reliability, which stems from their fuzzy or probabilistic interpretation of the underpinning logic languages. Probabilistic methods based on (subsets of) First Order Logic (FOL), for example, allow defining expressive knowledge but require expensive probabilistic inference routines \citep{li2023scallop,manhaeve2018deepproblog}. Analogously, fuzzy methods implement approximations of constraints and require a carefully informed design process \citep{badreddine2022ltn,vankrieken2022fuzzy}. Differently, methods based on propositional logics \citep{xu2018semantic,ahmed2022semantic,giunchiglia2024ccn} allow less expressive knowledge but benefit from tractable inference.
In this context, DL languages are a promising formalism for NeSy models since they are essentially subsets of FOL that benefit from more tractable inference routines while retaining enough expressiveness to formalize complex domain-specific knowledge \citep{vanharmelen2008kr,baader2003dlhandbook}.
NeSy approaches to DL are relatively under-explored when compared to other KR formalisms. Recent methods propose learning vectorial representations of ontologies \citep{chen2025ontologyembeddings}, approximate automated reasoning algorithms \citep{hitzler2025ddr}, or integrate the ontology within neural network classifiers as a form of regularization \citep{wu2022nesyfuzzyalc,slusarz2023ldl}. Understanding how they enhance the reliability of a model is difficult, since they generally evaluate the accuracy of the model in deductive reasoning or downstream tasks \citep{singh2025benchmarking}.

In this paper, we present NeSy method that combines a DL ontology with a neural network classifier by representing the ontology as a circuit \citep{choi2020pc,vergari2021compositional}. Circuits are compact computational representations of complex functions that are used, among other things, to represent boolean functions \citep{darwiche2002knowledge,bryant1986graph,darwiche2011sdd,darwiche2001decomposable}, complex probability distributions \citep{choi2020pc,loconte2024subtractive,darwiche2009modeling,anji2022scaling-pcs-lvd,gala2024pic}, and matrix factorizations \citep{loconte2025circuitfactorization}. They support tractable computation of complex transformations and queries \citep{vergari2021compositional,darwiche2002knowledge}, which enable, for instance, efficient probabilistic reasoning \citep{chavira2008wmc,de2007problog,riguzzi2013bundle}. Circuits have been used to propose different NeSy models \citep{xu2018semantic,ahmed2022semantic,manhaeve2018deepproblog,li2023scallop} where the circuits are produced using \textit{knowledge compilation} techniques \citep{darwiche2002knowledge}. Informally, a knowledge compiler takes as input a formula in propositional logic and outputs a circuit that allows tractable and efficient evaluation of the formula, preserving its logical meaning. \citet{rudolph2011dlfoundations} proposes to compile a DL ontology to a circuit as an intermediary representation to perform deductive reasoning using Datalog.

We adapt this approach and show that the resulting circuit can be used to:
\begin{itemize}
    \item[\textbf{C1}] generate a synthetic dataset that encapsulates the semantics of the ontology;
    \item[\textbf{C2}] efficiently perform deductive reasoning on a GPU;
    \item[\textbf{C3}] design NeSy classifiers that approximately or provably output predictions consistent with the background knowledge expressed in the ontology.
\end{itemize}

The paper is organized as follows: Section \ref{sec:dl-circuit} provides a general overview of our contributions. We introduce circuits and illustrate, by means of a simple example in the music domain, how the circuit compiled from an ontology can be used for data generation (\textbf{C1}), deductive reasoning (\textbf{C2}), and NeSy integration (\textbf{C3}). The following sections describe in more detail our contribution. In Section \ref{sec:background} we give an overview of the syntax and semantics of Description Logic, which serves as a basis for the knowledge compilation method described in Section \ref{sec:compilation}. In Section \ref{sec:related}, we review related methods that integrate ontologies with neural networks, as well as recent NeSy methods and their relationship with ontologies. In Section \ref{sec:experiments}, we experiment with the applications introduced in Section \ref{sec:dl-circuit}. In particular, in Section \ref{sec:data-generation} we show how to use the circuit as a data generation device (\textbf{C1}), which also serves as the basis for the remainder of the experiments. We then experiment by using the circuit as a deductive reasoner in Section \ref{sec:logical-reasoning} (\textbf{C2}) and compare its performance with existing DL reasoners. Finally, in Section \ref{sec:nesy}, we experiment using the circuit to obtain NeSy models and show that it produces reliable neural network models (\textbf{C3}). In Section \ref{sec:conclusion}, we summarize our contributions and highlight limitations and future directions.

\section{Tractable DL circuits}
\label{sec:dl-circuit}
We aim at designing a reliable NeSy model that combines a neural network model with an ontology encoding domain-specific knowledge. To do so, we need a practical computational approach to detect whether the predictions of the neural network are consistent with the ontology. Moreover, to achieve a seamless integration between the model and the ontology during the training phase, we need this approach to be fully differentiable with respect to the neural network's inputs. 

In this section, we show that compiling an ontology into a circuit achieves all our computational desiderata and, in addition, enables other useful applications beyond NeSy models, thanks to the circuit's properties.

Throughout the rest of the paper, we will refer to the same example scenario of automatically extracting the artistic influences of an artist. 

For the sake of concreteness, we assume we have access to a corpus of artists' biographies extracted from Wikipedia. For each entity mentioned in the corpus, we have some background knowledge available extracted from Wikidata, a humanly curated Knowledge Graph based on Wikipedia. We frame the problem as a link prediction task where we want to design a neural network model that predicts whether one artist has been influential to another artist, such as between \textit{Fugazi} and \textit{The Smiths} in the sentence Example \ref{ex:sentence}. 

\begin{figure}[htbp]
    \begin{minipage}{0.3\linewidth}
        \includegraphics[width=\textwidth]{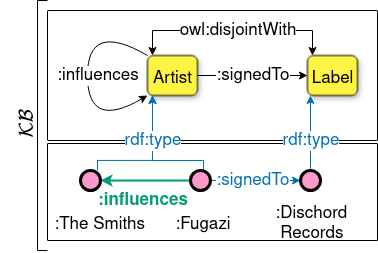}
        \caption{Knowledge Base $\KB$ represented in OWL2 using Graffoo. Yellow rectangles correspond to concepts, pink circles to individuals. Arrows between rectangles (resp. circles) represent constraints between concepts (resp. relations between individuals).}
        \label{fig:ex:kb}
    \end{minipage}%
    \hfill
    \begin{minipage}{0.65\linewidth}
        \includegraphics[width=\textwidth]{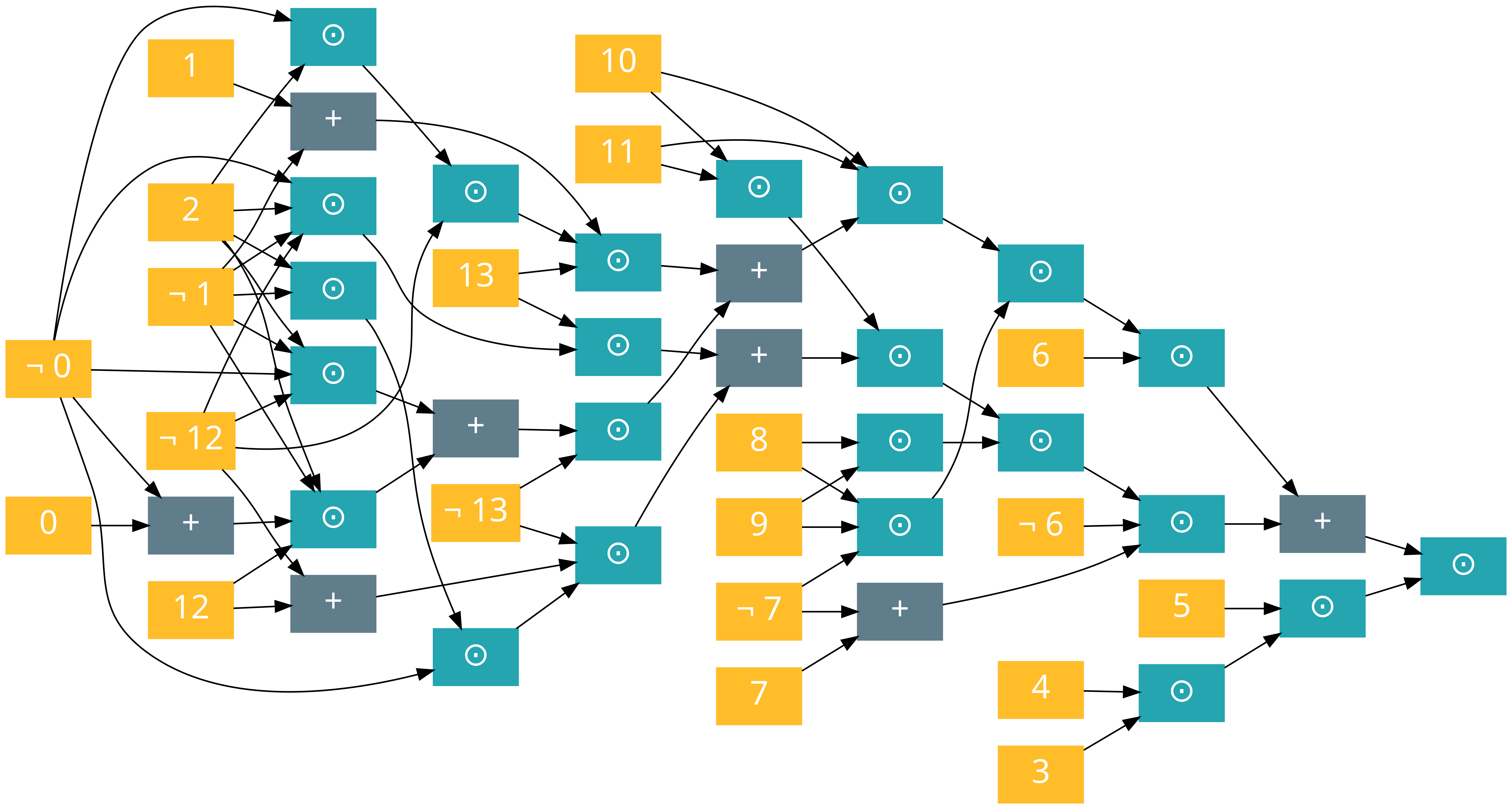}
        \caption{Circuit $\Co$. Input units (yellow) are labeled with their corresponding propositional variable and its corresponding negation. Each input corresponds to a part of the ontology (a concept, a role, or a role constraint). Sum units (grey) are labeled with $+$, product units (blue) with $\odot$.}
        \label{fig:ex:circuit}
    \end{minipage}
\end{figure}

For simplicity, we assume we can only observe two types of mutually disjoint entities in the corpus: artists (\eg \textit{Fugazi} and \textit{The Smiths}) and music labels (\eg \textit{Dischord Records}). Artists can be signed to music labels, but can only be influenced by other artists. This simple set of constraints represents our background knowledge and acts as an \textit{intensional} definition of the domain, represented in the form of an ontology $\ontology$ \citep{guarino2009ontology}. Figure \ref{fig:ex:kb} shows $\ontology$ using Graffoo\footnote{\url{https://essepuntato.it/graffoo/}}, an intuitive visual representation of OWL2, the standard ontology language developed within the KR and Semantic Web fields \citep{hitzler2009owl}. In Section \ref{sec:background}, we show how to formalize $\ontology$ using Description Logic. 

The purpose of $\ontology$ is to establish a domain-specific vocabulary (also called \textit{intensional knowledge}) that is then used to express relationships between entities observed in the real world (\textit{extensional knowledge}). 

In Example \ref{ex:sentence}, \textit{Fugazi} being an artist is a piece of extensional knowledge that can be represented by using $\ontology$. This is done in Figure \ref{fig:ex:kb} through the direct edge, labeled with \texttt{rdf:type}, between \textit{Fugazi} and the concept $\artistConcept$. Similarly, \textit{Fugazi} being signed to \textit{Dischord Records} is represented as a direct edge between their corresponding nodes, labeled with $\signedRole$. In this paper, we assume that all the extensional knowledge has been formalized in the form of a Knowledge Graph $\KG$ \citep{hogan2022kgbook}. In other words, $\KG$ contains a set of triples of the form $\langle \textit{subject}, \textit{relation}, \textit{object} \rangle$ between different entities.

In our link prediction task, the objective is to predict that there is an influence relation between \textit{Fugazi} and \textit{The Smiths}, represented as a green edge labeled with $\influeceRole$ in Figure \ref{fig:ex:kb}. Due to the lack of reliability discussed in Section \ref{sec:intro}, however, we might wrongly identify \textit{Dischord Records} as an artist, for instance, due to the ambiguity introduced by the sentence structure, and conclude that $\influeceRole$ holds between \textit{Dischord Records} and \textit{The Smiths} as well. 
Clearly, this is inconsistent with the knowledge expressed in the Knowledge Base $\KB = \ontology \cup \KG$ since \textit{Dischord Records} has been previously defined as a music label which cannot be an influence on an artist.
In this work, we show that by compiling the ontology $\ontology$ to a circuit $\Co$, one can detect and prevent these kinds of inconsistencies.

\begin{definition}[Circuit \citep{choi2020pc,vergari2021compositional}]
A circuit $c$ is a parameterized directed acyclic computational graph over variables $\X$ encoding a function $c(\X)$, and comprising three kinds of
computational units: input, product, and sum units. Each product or sum unit $n$ receives the outputs
of other units as inputs, denoted with the set $\cinput{n}$. Each unit $n$ encodes a function $c_n$ defined as: (i) $f_n(\cscope{n})$ if $n$ is an input unit, where $f_n$ is a function over variables $\cscope{n} \subseteq \X$, called its scope, (ii) $\prod_{j \in \cinput{n}} c_j(\cscope{j})$ if $n$ is a product unit, and (iii) $\sum_{j \in \cinput{n}} w_j c_j(\cscope{j})$ if $n$ is a sum unit, with $w \in \mathbb{R}$ denoting the weighted sum parameters.
\end{definition}

Figure \ref{fig:ex:circuit} shows a graphical representation of the circuit $\Co$ compiled using the method described in Section \ref{sec:compilation}, where inputs are ordered before outputs from left to right. Informally, each input to the circuit refers to a part of the ontology $\ontology$, such as the different concepts and roles defined and the relationships between them. A detailed description is provided in Section \ref{sec:compilation}.
By imposing different properties on the structure of a circuit, it is possible to guarantee tractable execution of complex queries and transformations \citep{choi2020pc,vergari2021compositional,darwiche2001decomposable}. In this paper, we rely on three main properties: smoothness, decomposability, and determinism.

\begin{definition}[Smoothness, decomposability, determinsm \citep{choi2020pc,vergari2021compositional,darwiche2002knowledge}]
    A circuit $c$ is smooth if for every sum unit $n$, its input units depend all on the same variables, \ie $\forall i, j \in \cinput{n}: \cscope{i} = \cscope{j}$. A circuit $c$ is decomposable if the inputs of every product unit $n$ depend on disjoint sets of variables, \ie $\forall i, j \in \cinput{n}, i \neq j: \cscope{i} \cap \cscope{j} = \emptyset$. A circuit $c$ is deterministic if for every sum unit $n$, its inputs have disjoint supports, i.e., $\forall i, j \in \cinput{n}, i \neq j : \csupport{i} \cap \csupport{j} = \emptyset$. The support of a sum unit $n$ is the set of assignments to its scope such that the output of $n$ is non-zero, i.e., $\csupport{n} = \{ x \in \mathsf{val}(\X)\ \vert\ n(x) \neq 0 \}$.
\end{definition}

Smooth and decomposable circuits support tractable marginalization \ie, it is possible to \textit{ignore} some variables when evaluating the encoded function, which is a fundamental requirement for modeling complex probability distributions \citep{poon2011sum,peharz2015theoretical} that is generally hard \citep{broadrick2025marginalization}. Similarly, they support complex information-theoretical queries, such as the similarity between two probability distributions through their KL-divergence \citep{vergari2021compositional}. Similarly, decomposable and deterministic circuits support boolean \citep{darwiche2002knowledge} and probabilistic \citep{chavira2008wmc,darwiche2009modeling} reasoning as well as complex queries that are traditionally intractable, such as Maximum A Posteriori (MAP) \citep{choi2020pc,darwiche2009modeling}.
These structural properties are usually enforced by carefully designing the circuit construction phase \citep{poon2011sum,dennis2012learning,gens2013learning,peharz2020random,loconte2025circuitfactorization} or by using knowledge compilation techniques \citep{darwiche2002knowledge,kisa2014probabilistic,dang2021strudel}. The circuit of Figure \ref{fig:ex:circuit}, obtained through knowledge compilation, is smooth, decomposable, and deterministic.

\paragraph{$\Co$ for (probabilistic) deductive reasoning (\textbf{C2})}
In Section \ref{sec:compilation} we describe how $\Co$ faithfully encodes the semantics of $\ontology$, \ie $\Co(\mathbf{x}) = 1$ if and only if $\mathbf{x}$ corresponds to some knowledge assertion consistent with $\ontology$, otherwise $\Co(\mathbf{x}) = 0$. In other words, $\csupport{\Co}$ is the set of assignments corresponding to consistent knowledge with respect to $\ontology$.
This means that we can perform deductive reasoning by parameterizing $\Co$ as a boolean circuit \citep{darwiche2002knowledge,peharz2015theoretical} (input units are indicator functions and sum units have unitary weights) and evaluate the consistency of an assertion as $\Co(\mathbf{x})$.
Consider the case in which we want to check the consistency of \textit{Dischord Records} being an influence for \textit{The Smiths} given $\KG$. For the sake of simplicity, we present here a simplified encoding of $\mathbf{x}$ that follows the same principles as the one detailed in Equation \ref{eq:cbf} of Section \ref{sec:compilation}: we construct $\mathbf{x}$ as the concatenation of $\mathbf{s}$, $\mathbf{o}$ and $\mathbf{r}$ defined as:
{
\begin{gather}
    \nonumber \\
    \mathbf{s} = [\eqnmark{sartist}{0}, \eqnmark{slabel}{1}] \quad \quad \mathbf{r} = [\eqnmark{rinfluences}{1}, \eqnmark{rsignedto}{0}] \quad \quad [\eqnmark{oartist}{1}, \eqnmark{olabel}{0}]
    \label{eq:informal-domino-encoding}
\end{gather}
\annotate{below,left}{sartist}{\artistConcept}
\annotate{below,right}{slabel}{\labelConcept}
\annotate{above,left,label below}{rinfluences}{\influeceRole}
\annotate{above,right,label below}{rsignedto}{\signedRole}
\annotate{below,left}{oartist}{\artistConcept}
\annotate{below,right}{olabel}{\labelConcept}
}

Informally, $\mathbf{s}$, $\mathbf{o}$, and $\mathbf{r}$ are multi-hot encodings of the knowledge contained in $\KG$ about \textit{Dischord Records} and \textit{The Smiths} where each element corresponds to one input of $\Co$. We have that $\Co(\mathbf{x}) = 0$, since the assertion is not consistent with $\ontology$. In Section \ref{sec:logical-reasoning}, we implement this idea by relying on vectorized circuit implementation, which can be efficiently evaluated on a GPU \citep{peharz2020einsum,liu2024pyjuice,maene2025klay,loconte2025circuitfactorization}. We demonstrate that our reasoner is up to three orders of magnitude faster when used with a large amount of data.

By changing the parameterization of $\Co$ to retrieve its probabilistic semantics (input units are probability distributions and sums' weights are non-negative and sum up to one), it is trivial to obtain a probabilistic deductive reasoner. Informally, we have that $\Co$ will encode a probability distribution over the assertions that are consistent with $\ontology$, where $\Co(\mathbf{x}) = p \in [0, 1]$ is the \textit{belief} that the assertion encoded by $\mathbf{x}$ has been observed. Crucially, $\Co(\mathbf{x}) > 0$ only for assertions that are consistent with the ontology $\ontology$.

For example, assume the link prediction model provides some probabilistic confidence score over its predictions, for example
\begin{equation*}
    Pr(\text{``Dischord Records is an influence for The Smiths''}) = 0.34
\end{equation*}

We can parameterize the input unit related to $\influeceRole$ as a Bernoulli distribution with parameter $p = 0.34$ and parameterize the rest according to the knowledge in $\KG$ \eg the input units related to \textit{Dischords Records} being classified as $\labelConcept$ are parameterized by $Bernoulli(1)$ (or, equivalently, an indicator function). Assuming a uniform belief over all the possible assignments consistent with $\ontology$, implemented by parameterizing the sum units with uniformly distributed weights, we can compute the probability that the information extracted is consistent as $\Co(\mathbf{x}) = 0.66$. It is possible to update the belief over the assignments by changing the parameterization of the sum units, effectively conditioning the probability distribution encoded by $\Co$ \citep{shao2022conditionalgatefunction}.

\paragraph{$\Co$ for synthetic data generation (\textbf{C1})}
Alongside probabilistic deductive reasoning, the probabilistic semantics of $\Co$ can be used to sample from the circuit $\Co$ by leveraging smoothness and decomposability \citep{choi2020pc,lang2022pc-sampling}. A sample $\mathbf{\hat{x}}$ drawn from $\Co$, $\mathbf{\hat{x}} \sim \Co$, encodes an assertion that can appear in $\KG$, following the structure of Equation \ref{eq:informal-domino-encoding}. More importantly, it corresponds to knowledge consistent with $\ontology$, \ie $\Co(\mathbf{\hat{x}}) > 0$. Note that when sampling from $\Co$, $\mathbf{\hat{x}}$ does not encode information about concrete entities (e.g. \textit{Fugazi} or \textit{The Smiths}) but rather it specifies the intensional description of an entity. In other words, $\mathbf{\hat{x}}$ is a \textit{knowledge configuration} describing a hypothetical entity, consistent with $\ontology$, that could be observed in the real world. In Section \ref{sec:data-generation}, we show how this approach can be used to produce a synthetic dataset that reflects the semantics expressed in the ontology.

\paragraph{$\Co$ for NeSy models (\textbf{C3})}
Alongside their properties, one key aspect of circuits is their computational graph nature, which benefits from a clear feed-forward semantics. This can be seen from the circuit in Figure \ref{fig:ex:circuit}: the input assignment is processed following the direction of the computational graph, combining values through sums and products until an output is obtained. As a consequence, their output is fully differentiable with respect to their inputs \citep{peharz2015theoretical}, enabling a natural integration with the learning process of neural networks. Recent works in the NeSy field have shown that it is possible to leverage this aspect to approximately \citep{xu2018semantic} or provably \citep{ahmed2022semantic} obtain neural-network predictions that are consistent with user-specified constraints \citep{calanzone2025nesylm,ahmed2023pseudo}. These approaches are built on the idea of compiling constraints to circuits using knowledge compilation techniques \citep{darwiche2002knowledge} and interpreting the neural network predictions as a parameterization of the resulting circuit. 

These arguments directly apply to the circuit $\Co$. In Section \ref{sec:experiments}, we show how these NeSy solutions allow a neural network to produce predictions in the form of Equation \ref{eq:informal-domino-encoding} that are consistent with the ontology $\ontology$. In particular, we will experiment with Semantic Probabilistic Layers (SPL, Equation \ref{eq:spl}) \citep{ahmed2022semantic}, where the circuit $\Co$ is parameterized by a neural network and its structural properties are used to extract a consistent prediction. We also experiment with Semantic Losses (SL, Equation \ref{eq:sl}) \citep{xu2018semantic}, where the probabilistic reasoning capabilities of the circuit are used as a regularization term in the objective function optimized when training the neural network. Both approaches allow effective NeSy integrations with different guarantees: on the one hand, an SPL implements \textit{hard constraints} and guarantees consistent predictions, while a SL approach implements \textit{soft constraints}, tolerating situations in which the knowledge of $\ontology$ is imprecise. In Section \ref{sec:nesy} we experiment with $\Co$ using both approaches and show that they achieve competitive or better performances when compared to neural baselines and DeepProbLog \citep{manhaeve2018deepproblog} while being more reliable. Moreover, we show that our NeSy methods support a straightforward integration of background knowledge.

\section{Background}
\label{sec:background}
In this section, we provide an overview of Description Logic (Section \ref{sec:dl}) by introducing its motivations, syntax, semantics, and applications. For a more detailed treatment, we refer the reader to \citet{baader2003dlhandbook}. We then provide an overview of related works that combine ontologies and neural networks and review NeSy approaches that can be used to integrate ontology with neural network models (Section \ref{sec:related}). Finally, in Section \ref{sec:compilation}, we provide a detailed explanation of the compilation approach of \citet{rudolph2012dlobdd} and show how it is used to obtain the circuit $\Co$. Throughout the whole chapter, we will refer to $\Co$, $\KB$, $\KG$, and $\ontology$ from the example of Figures \ref{fig:ex:kb} and \ref{fig:ex:circuit} if not explicitly stated.

\subsection{Description Logic}
\label{sec:dl}
Description Logics (DL) are formal languages used to define, manipulate, and reason about domain-specific knowledge. A DL knowledge base $\KB$ is defined as the combination of \textit{terminological} and \textit{assertional} knowledge. The terminological knowledge establishes the domain-specific vocabulary by defining a set of concepts $\concepts$, a set of roles (binary relations) $\roles$ between the concepts $\concepts$, and complex descriptions obtained by combining concepts and roles through different constructs. Assertional knowledge instantiates named individuals and their relations using the concepts $\concepts$, relations $\roles$, and complex combinations of them \citep{baader2003dlhandbook}. In literature, terminological knowledge is notated as $TBox$ and assertional knowledge as $ABox$. In this paper, we refer to them as, respectively, ontology $\ontology$ and Knowledge Graph $\KG$. 

For example, $\artistConcept \in \concepts$ and $\influeceRole \in \roles$ are \textit{statements} (or \textit{axioms}) in $\ontology$, while $\artistConcept(\textit{Fugazi})$ and $\signedRole(\textit{Fugazi}, \textit{Dischord Records})$ are \textit{assertions} contained in $\KG$. Assertions can be created manually or automatically, such as by using the link prediction model hypothesized in Section \ref{sec:dl-circuit} with $\influeceRole(\textit{Fugazi}, \textit{The Smiths})$ represented in green in Figure \ref{fig:ex:kb}.
Complex concept descriptions are obtained by combining concepts $\concepts$ and roles $\roles$ using \textit{concept constructors}, \eg boolean connectives, such as defining the set of individuals that is \textit{not} an artist, but does not have to be a music label.

Different choices of concept constructors identify different DL \textit{fragments}. Each fragment exhibits a different tradeoff between expressivity and computational complexity in automated reasoning tasks. For example, the $\mathcal{SROIQ}$ fragment, underpinning OWL ontologies\footnote{\url{https://www.w3.org/TR/owl2-profiles/}}, is a decidable subset of First Order Logic with \texttt{NExpTime-hard} complexity. Less expressive but more tractable fragments have been proposed in literature by carefully selecting and analyzing concept constructors\footnote{See \url{http://www.cs.man.ac.uk/~ezolin/dl/} for an extensive review of DL fragments and their complexity.} such as $\ALC$ \citep{schmidtschauss1992alc}, $\mathcal{EL}$ \citep{baader2008el}, or DL-lite \citep{calvanese2005dllite}.
The $\ALC{}$ fragment is often considered the \textit{default} DL fragment since it allows tractable reasoning in polynomial space complexity while still retaining enough representational power to model complex scenarios \citep{rudolph2011dlfoundations}. In the next sections we overview the syntax and semantics of the $\ALCI{}$ fragment, which we will compile to a circuit in Section \ref{sec:compilation}.

\subsubsection{Syntax}
\begin{table}[ht]
    \centering
    \begin{tabular}{ccc}
    \toprule
    & Syntax & Semantics \\ \midrule
    atomic role & $R$ & $\RI \subseteq \DeltaI \times \DeltaI$ \\
    inverse role & $R^{-}$ & $\{ \braI \mid \arbI \in \RI \}$ \\
    role inclusion & $R \sqsubseteq S$ & $\RI \subseteq S^{\I}$ \\
    
    atomic concept & $A$ & $\AI \subseteq \DeltaI$ \\
    intersection & $A \sqcap B$ & $\AI \cap \BI$ \\
    union & $A \sqcup B$ & $\AI \cup \BI$ \\
    complement & $\lnot A$ & $\DeltaI \setminus \AI$ \\
    top concept & $\top$ & $\DeltaI$ \\
    bottom concept & $\bot$ & $\emptyset$ \\
    existential restriction & $\exists R.B$ & $\{ a \in \DeltaI \mid \exists b.\arb \in \RI \land b \in \BI \}$ \\
    universal restriction & $\forall R.B$ & $\{ a \in \DeltaI \mid \forall b.\arb \in \RI \rightarrow b \in \BI \}$ \\ 
    concept inclusion & $A \sqsubseteq B$ & $\AI \subseteq \BI$ \\
    concept equivalence & $A \equiv B$ & $\AI = \BI$ \\ \midrule

    individual & $a$ & $\aI \in \DeltaI$ \\ 
    concept assertion & $A(a)$ & $\aI \in \AI$ \\
    role assertion & $R(a, b)$ & $\arbI \in \RI$ \\ \bottomrule
    \end{tabular}
    \vspace{1em}
    \caption{\ALC{} syntax and semantics where $a, b$ are named individuals, $A, B$ concepts in \concepts{}, $R, S$ roles in \roles{}. The top half refers to constructors used in the \tbox{} while the bottom half refers to constructors used in the \abox{}.}
    \label{tab:alc-syntax-semantics}
\end{table}

The $\ALCI{}$ language is an extension of the $\ALC{}$ language (Attributive Language with Complement), introduced as a minimal tractable language of practical interest \citep{schmidtschauss1992alc}. It defines the constructors listed in Table \ref{tab:alc-syntax-semantics}, which can be composed together to construct complex descriptions. The $\ALCI{}$ fragment extends $\ALC$ by introducing role inverses (included in Table \ref{tab:alc-syntax-semantics}), without introducing additional complexity.

Even though the $\ALCI{}$ DL is a tractable fragment, it can be used to represent some of the most used constructs defined in the OWL2 language: 
\begin{itemize}
    \item concept subsumption, \ie hierarchically organize the concepts $\concepts$;
    \item concept disjunction, \ie defining mutual exclusivity between the concepts $\concepts$;
    \item domain and range constraints, \ie defining constraints on the classification of entities that can be used as subjects and objects of a role assertions.
\end{itemize}

For example, the ontology $\ontology$ of Figure \ref{fig:ex:kb} can be represented in DL as

\begin{align}
    \artistConcept &\sqsubseteq \lnot \labelConcept & \longrightarrow & \textit{``A artist is not a music label.''} \label{ex:o-1} \\
    \top &\sqsubseteq \forall \influeceRole.\artistConcept & \longrightarrow & \textit{``Only artists can be the object of an influence assertion.''} \label{ex:o-2} \\
    \top &\sqsubseteq \forall \influeceRole^-.\artistConcept & \longrightarrow & \textit{``Only artists can be the subject of an influence assertion.''} \label{ex:o-3} \\
    \top &\sqsubseteq \forall \signedRole.\labelConcept & \longrightarrow & \textit{``Only labels can be the object of a signing assertion.''} \label{ex:o-4} \\
    \top &\sqsubseteq \forall \signedRole^-.\artistConcept & \longrightarrow & \textit{``Only artists can be the subject of a signing assertion.''} \label{ex:o-5}
\end{align}

where $\concepts = \{ \artistConcept, \labelConcept \}$, and $\roles = \{ \influeceRole, \signedRole \}$. Statement \eqref{ex:o-1} is an example of a complex concept description. Note that $X \sqsubseteq Y$ is equivalent to $\top \sqsubseteq \lnot A \sqcup B$ and the usual boolean properties on conjunctions and disjunctions also apply to DL constructors. For instance, $\artistConcept \sqsubseteq \lnot \labelConcept$ can be equivalently written as $\top \sqsubseteq \lnot (\artistConcept \sqcap \labelConcept)$ or $\top \sqsubseteq (\lnot \artistConcept \sqcup \lnot \labelConcept)$.

The constructors of Table \ref{tab:alc-syntax-semantics} allow the definition of more complex concepts than the one we show in Figure \ref{fig:ex:kb}. For instance, an artist that is not influenced by any other artist can be formalized in DL as
\begin{equation*}
    \mathsf{HermitArtist} \sqsubseteq (\artistConcept \sqcap \forall \influeceRole.\bot)
\end{equation*}

Using the constructors of Table \ref{tab:alc-syntax-semantics}, the $\KG$ of Figure \ref{fig:ex:kb} is represented as

\begin{align}
    &\artistConcept(\textit{Fugazi}) \quad \artistConcept(\textit{The Smiths}) & \longrightarrow & \textit{``Fugazi and The Smiths are artists.''} \label{ex:kg-1} \\
    &\labelConcept(\textit{Dischord Records}) & \longrightarrow & \textit{``Dischord Records is a music label.''} \label{ex:kg-2} \\
    &\signedRole(\textit{Fugazi}, \textit{Dischord Records}) & \longrightarrow & \textit{``Fugazi is signed to Dischord Records.''} \label{ex:kg-3} \\
    &\influeceRole(\textit{Fugazi}, \textit{The Smith}) & \longrightarrow & \textit{``Fugazi influenced The Smith.''} \label{ex:kg-4}
\end{align}

As discussed in Section \ref{sec:dl-circuit}, the output of the link prediction model must be consistent with $\KB$ in order to be integrated within $\KG$, otherwise one might generate contradicting knowledge that cannot be trusted. This notion of consistency is central to automated reasoning algorithms and is based on the DL semantics, \ie the interpretation that should be used to give meaning to the DL constructs.

\subsubsection{Semantics}
The semantics of $\ALCI{}$ is defined by an interpretation $\I = (\DeltaI, \cdot^{\I})$ where $\DeltaI$ is a non-empty set called the \textit{domain of interpretations} and $\cdot^{\I}$ is an \textit{interpretation function}, which assigns atomic concepts to subsets of $\DeltaI$, atomic roles to elements of $\DeltaI \times \DeltaI$ and individuals to elements of $\DeltaI$, where different named individuals refer to different elements of $\DeltaI$ (also called unique name interpretation \citep{baader2003dlhandbook}). The interpretation function is extended to the concept constructors as shown in Table \ref{tab:alc-syntax-semantics}.

For example, an interpretation $\mathcal{I}$ for $\KG$ is
\begin{align}
    \textit{Fugazi}^{\I} \in \artistConcept^{\I} \quad \textit{The Smiths}^{\I} \in \artistConcept^{\I} \quad \textit{Dischord Records}^{\I} \in \labelConcept^{\I} &\longrightarrow \text{ From \eqref{ex:kg-1} and \eqref{ex:kg-2}} \\
    \langle \textit{Fugazi}^\I, \textit{DischordRecords}^\I \rangle \in \signedRole^{\I} &\longrightarrow \text{ From \eqref{ex:kg-3}} \\
    \langle \textit{Fugazi}^\I, \textit{The Smiths}^\I \rangle \in \influeceRole^{\I} &\longrightarrow \text{ From \eqref{ex:kg-4}}
\end{align}

If we integrate the statement $\influeceRole(\textit{Dischord Records}, \textit{The Smiths})$ in $\KG$, then the interpretation $\mathcal{I}$ is extended with 
\begin{align}
    \langle \textit{Dischord Records}^\I, \textit{The Smiths}^\I \rangle \in \influeceRole^{\I}
\end{align}

Because of the definitions in $\ontology$ and the semantics of Table \ref{tab:alc-syntax-semantics}, $\I$ must also include $\textit{Dischord Records}^{\I} \in \artistConcept^{\I}$, since it is the object of an $\influeceRole$ assertion (Equation \eqref{ex:o-3}). Given the ontology $\ontology$, $\mathcal{I}$ is not a legal interpretation anymore, since $\textit{Dischord Records}^{\I} \in \artistConcept^{\I}$ and $\textit{Dischord Records}^{\I} \in \labelConcept^{\I}$ are \textit{contradicting} assertions due to Equation \eqref{ex:o-1}. The concept of contradictions is central to the automated reasoning procedures in DL, as we explain in the next section.

\subsubsection{Reasoning}
\label{sec:reasoning}
Formally, an interpretation $\I$ is consistent with a $\KB$ if $\I$ \textit{satisfies} the concept $C$, written $\I \models C$, for all $C \in \concepts$. We say that $\I \models C$ if $\CI$ is not empty. A $\KB$ is consistent if all of its concepts are consistent for any interpretation $\I$. 
Consistency is central to automated reasoning since most of the interesting queries on $\KB$ (\eg classification, subsumption test) can be reduced to consistency checking \citep{baader2003dlhandbook}. In other words, an efficient algorithm for consistency checking enables a multitude of efficient automated reasoning tasks. Moreover, the complexity of DL fragments is measured as the complexity of the algorithm used to perform consistency checking.

Depending on the fragment at hand, different algorithms have been proposed in literature. The most common approaches are based on the tableau algorithm or its variations, which frame the consistency decision problem as a search algorithm. Statements and assertions of $\KB$ are iteratively processed until an inconsistency is generated. If the procedure ends without conflicts, $\KB$ is satisfiable. Popular DL reasoners such as Pellet \citep{sirin2007pellet}, HermiT \citep{glimm2014hermit}, and Konclude \citep{steigmiller2014konclude} rely on carefully designed heuristics to efficiently explore the huge search space. Other approaches include leveraging the structural properties of specific fragments \citep{yazakov2012elk,amir2017scalablereasoning} or reducing DL reasoning to other logic languages \citep{eiter2012rewriting,hustadt2004datalogdl,lukacsy2009dlog,grosof2003dlprograms,gaggl2016aspdl}.

\section{Related works}
\label{sec:related}
Different approaches have been proposed in recent years to combine DL ontologies with neural networks. These mainly follow two lines of research: deep deductive reasoners \citep{hitzler2025ddr} and ontology embeddings \citep{chen2025ontologyembeddings}.

\paragraph*{Deep Deductive Reasoners}
Deep deductive reasoners approximate deductive reasoning using neural networks by learning a representation of an ontology and its elements to derive sound and meaningful conclusions \citep{hitzler2025ddr}. This is generally done to understand whether different architectural choices, designed after DL semantics, allow neural networks to perform deductive reasoning in an efficient (as compared to traditional reasoners) yet reliable way without the use of NeSy methods \citep{hohenecker2020ddr,zhu2023ddr,eberhart2020ddr,ebrahimi2021ddr,makni2019ddr}. A different approach has been proposed by \citet{bianchi2019ltnddr}, where Logic Tensor Networks \citep{badreddine2022ltn}, a prominent neuro-symbolic method based on differentiable fuzzy FOL, are used to approximate the reasoning process. \citet{slusarz2023ldl} and \citet{wu2022nesyfuzzyalc} also rely on differentiable fuzzy logic to obtain some level of deductive reasoning, but rely on the fuzzy logic semantics of DL \citep{straccia1998fuzzydl,borgwardt2017fuzzydl}.

\paragraph*{Ontology Embeddings}
Ontology embeddings, on the other hand, compute vectorial representations of the elements of $\ontology$ that can be specialized for additional tasks beyond deductive reasoning, such as classification or ontology alignment \citep{chen2025ontologyembeddings}. These include methods inspired by the word2vec \citep{mikolov2013word2vec} approach, where the OWL representation of $\ontology$ is used to synthesize sentences from the ontology that are then embedded as vectors \citep{chen2021owl2vec,he2022bertmap,gosselin2023sorbet}. These approaches are closely related to KG embeddings approaches where an ontology is used as a backbone to produce training data a KG \citep{ristoski2019rdf2vec} or to influence the training process of general KG embedding methods \citep{damato2021transowl,hubert2023enhancing}. Other approaches approach the problem by searching for geometric translations that capture parts of DL semantics \citep{gutierrezbasulto2018ontologyembeddings}, such as using spheres \citep{kulmanov2019sphericalembeddings}, boxes \citep{xiong2022box,yang2025boxes,jackermeier2024box}, or cones \citep{ozcep2020cone,ozcep2023cone}.

\paragraph*{Knowledge compilation and circuits in the DL landscape}
While our approach might seem similar to the ones described here, it fundamentally differs in its spirit. Deep deductive reasoners and ontology embeddings learn and/or apply (possibly latent) inference rules to derive conclusions using vectorial representations instead of symbolic ones. The circuit $\Co$ introduced in Section \ref{sec:dl-circuit} acts instead as a compact representation of the knowledge that can be expressed by $\ontology$. Whether this is used to obtain deductive reasoning or a representation of the elements in the ontology depends on the way $\Co$ is queried and transformed. The flexibility offered by circuits can be considered as an ensemble of efficient computational routines that one can use to design advanced DL systems, such as implementing different probabilistic \citep{lukasiewicz2008probabilisticdl,damato2008bayesiandl,botha2020balc} or fuzzy \citep{straccia1998fuzzydl,straccia2001reasoningfuzzydl,bobillo2009fuzzydl} DL semantics or combinations of them \citep{lukasiewicz2009fuzzyprobabilisticdl}.

The idea of encoding the semantics of an ontology in a compact representation is a form of knowledge compilation \citep{darwiche2002knowledge}. \citet{rudolph2012dlobdd} explores this approach by constructing and refining a propositional formula that encodes the ontology's set of models by compiling it to an OBDD circuit \citep{bryant1986graph}. We give a more detailed description of this method in Section \ref{sec:compilation}. A similar approach has been proposed by \citet{furbach2009linklessdl}, where a compact encoding that can be tractably queried has been proposed. The method, however, does not produce a circuit and hence does not trivially benefit from circuits' properties. Related approaches that rely on a circuit's properties have also been explored, including implementing the tableau algorithm using operations on circuits \citep{gore2014bddtab} and probabilistic deductive reasoners \citep{riguzzi2013bundle}.

\paragraph{NeSy literature in the context of DL}
In this paper, we rely on the properties of circuits to design a NeSy models that integrate an ontology within a neural network using Semantic Losses \citep{xu2018semantic} and Semantic Probabilistic Layers \citep{ahmed2022semantic}, as briefly introduced in Section \ref{sec:dl-circuit}. Note, however, that other methods have been proposed in recent years to integrate neural networks with user-defined constraints \citep{giunchiglia2022logicconstraints,morettin2021logic}, which could potentially serve a similar purpose, albeit without the properties guaranteed by circuits.
For instance, it is possible to reduce DL semantics to Answer Set Programming (ASP) \citep{gaggl2016aspdl} and rely on NeSy ASP methods such as NeurASP \citep{yang2020neurasp} or dPASP \citep{geh2024dpasp}; reduce DL to Prolog \citep{lukacsy2009dlog} and use DeepProbLog-based approaches \citep{manhaeve2018deepproblog,winters2021deepstochlog,desmet2023deepseaproblog} or reduce DL to DataLog \citep{grosof2003dlp,eiter2012rewriting,hustadt2004datalogdl} and use differentiable DataLog reasoners such as Scallop \citep{huang2021scallop,li2023scallop}. We note, however, that the tractability of a DL fragment is not always preserved in these reductions. Some tractable DL fragments, for instance, become undecidable when interpreted using ASP semantics \citep{distefano2024dlfinitemodels}.

\section{Compile a DL ontology to a circuit}
\label{sec:compilation}
In this section, we provide a detailed description of the compilation algorithm proposed by \citet{rudolph2012dlobdd} where a $\mathcal{SHIQ}b$ ontology is compiled to an OBDD circuit \citep{bryant1986graph}. Differently from the original work, we focus on the $\ALCI$ DL and compile the ontology to an SDD circuit \citep{darwiche2011sdd}, a generalization of OBDDs whose representation can be exponentially more compact than OBDDs \citep{bova2016sddsuccint}. In other words, the circuit $\Co$ shown in Figure \ref{fig:ex:circuit} might contain exponentially less nodes than the corresponding OBDD circuit. This directly impacts the efficiency of evaluating the computational graph implemented by the circuit and, as a consequence, the efficiency of the training and inference phase in our NeSy models.

\citet{rudolph2012dlobdd} present a reduction from $\ontology$ to a propositional logic formula through the use of a \textit{domino} metaphor. A domino interpretation (or simply, a domino) is a triple $\domino$ where $\mathcal{A}$ and $\mathcal{B}$ are sets of (complex) concepts while $\mathcal{R}$ is a set of roles.
Given an interpretation $\I$ and a set $\mathcal{C} \subseteq \concepts$ of (complex) concepts, the \textit{domino projection} of $\I$ \wrt $\mathcal{C}$ is the set that contains the triple $\domino$ such that
\begin{equation*}
    \mathcal{A} = \{C \in \mathcal{C} \mid \delta_1 \in C^{\I} \} \quad \mathcal{R} = \{ R \in \mathbf{R} \mid (\delta_1, \delta_2) \in R^{\I} \} \quad \mathcal{B} = \{ C \in \mathcal{C} \mid \delta_2 \in C^{\I} \}
\end{equation*}
for all $\delta_1, \delta_2 \in \DeltaI$, $\mathcal{A}, \mathcal{B} \subseteq \mathcal{C}$ and $\mathcal{R} \subseteq \roles$.

In other words, the domino projection of $\I$ \wrt $\mathcal{C}$ encodes which roles between two individuals $\delta_1$ and $\delta_2$ are consistent \wrt $\ontology$ if $\delta_1$ and $\delta_2$ are classified by the elements of $\mathcal{A}$ and $\mathcal{B}$, respectively. A domino projection does not faithfully represent the structure of $\I$, however, it allows the reconstruction of a model $\I$ such that $\I \models \ontology$. This property is satisfied if $\mathcal{C}$ is chosen to be the set of \textit{parts} of $\ontology$, written as $\parts(\KB)$. Essentially, a part of $\ontology$ is either an atomic concept or a role restriction on an atomic concept. \citet{rudolph2012dlobdd} describes an algorithm to extract parts from any $\ALCI$ ontology in polynomial time.

For example, the parts of the $\KB$ of Figure \ref{fig:ex:kb} are
\begin{align*}
    \parts(\KB) = \{ \artistConcept, \labelConcept, \forall \influeceRole.\artistConcept, \forall \influeceRole^-.\artistConcept, \forall \signedRole.\labelConcept, \forall \signedRole^-.\artistConcept \}
\end{align*}

Given a set of dominoes $\mathbb{D}$, it is possible to construct an interpretation $\canonicaldomino^\I$ from it, called the \textit{canonical domino interpretation}, such that $\canonicaldomino^\I \models \ontology$ \citep{rudolph2012dlobdd}. Informally, this is done by first collecting the \textit{biggest} set of dominoes $\mathbb{D}_0$ that includes all the models of $\ontology$ as well as interpretations that are not models of $\ontology$. The set $\mathbb{D}_0$ is then iteratively refined until only models of $\KB$ can be constructed from it. Note that the set $\mathbb{D}_0$ is exponential in the size of the $\ontology$ and is hence impractical to be represented explicitly. By relying on knowledge compilation, it is possible to compactly encode it as a boolean formula using characteristic boolean functions \citep{bryant1992obddmanipulation}. Moreover, it is possible to perform the refinement steps efficiently through the use of tractable transformation queries supported by the circuit, until $\canonicaldomino$ is obtained.

Figure \ref{fig:ex:circuit} shows the circuit obtained by compiling $\ontology$ as defined in Equations \eqref{ex:o-1}-\eqref{ex:o-5}. The compilation algorithm is summarized into two main steps:
\begin{enumerate}
    \item Normalize $\ontology$ to Negative Normal Form (NNF);
    \item Compile a boolean formula encoding normalized $\ontology$ into a tractable circuit.
\end{enumerate}

The NNF normalization phase leverages syntactic and semantic equivalences to rewrite $\ontology$ so that negation only appears before an atomic concept. Suppose for example, that we have a statement describing Punk artists as artists that can not be influenced by neither Pop nor Classical artists. The statement and its equivalent NNF translation is
\begin{equation*}
    \mathsf{Punk} \sqsubseteq \forall \influeceRole^{-1}.\lnot (\mathsf{Pop} \sqcup \mathsf{Classical}) \quad\underset{\textit{NNF}}{\longrightarrow}\quad \mathsf{Punk} \sqsubseteq \forall \influeceRole^{-1}.(\lnot \mathsf{Pop} \sqcap \lnot \mathsf{Classical})
\end{equation*}

Additionally, $\ontology$ is \textit{flattened} by rewriting existential and universal restrictions so that they are only asserted on atomic concepts. For example, the previous NNF axiom is replaced by
\begin{align*}
     \mathsf{Punk} \sqsubseteq \forall \influeceRole^{-1}.X \quad \top \sqsubseteq \lnot X \sqcup (\lnot \mathsf{Pop} \sqcap \lnot \mathsf{Classical})
\end{align*}
where $X$ is a fresh concept name. From here on, we assume that $\ontology$ has been normalized to NNF and flattened if not explicitly stated.

The boolean formula that is compiled to $\Co$ is defined on the set of variables $\Vars = \{ v_X : X \in \roles \cup (\parts(\ontology) \times \{ 1, 2 \})$ where $v_X$ is a fresh variable that refers to $X$. Intuitively, $\Vars$ contains one variable for each element that could be in the sets $\mathcal{A}$, $\mathcal{B}$ (\eg $(\artistConcept, 1)$ and $(\artistConcept, 2)$) or $\mathcal{R}$ of a domino $\domino$. 

This can also be seen in Figure \ref{fig:ex:circuit}: the circuit has 14 variables, two for each element in $|\parts{(\ontology)}| = 6$ and one for each role.
Differently from the approach of \citet{rudolph2012dlobdd}, we encode $\mathbb{D}_0$ as a Conjunctive Normal Form (CNF) formula. This allows us to leverage the more optimized knowledge compilers that rely on heuristics to compile CNF formulas~\citep{oztok2015top-down-sdds,oztok2017compiling}. Moreover, we assert additional clauses on the CNF that allow us to compile an initial domino set $\mathbb{D}_0$ where some refinements are preemptively performed. For instance, we constrain the CNF such that it only contains dominoes that do not violate universal restrictions.

\begin{algorithm}
    \caption{Compute the circuit $\cbf{\mathbb{D}_0}$ encoding the domino set $\mathbb{D}_0$}
    \label{alg:d0}
    \begin{algorithmic}
    \STATE $\phi^{\text{kb}} := \bigwedge\limits_{A \in \tbox{}} \cbf{A, 1} \land \cbf{A, 2}$
    \STATE $\phi^{\text{uni}} := \bigwedge\limits_{\forall R.A \in \parts(\tbox{})} \cbf{\forall R.A, 1} \land \cbf{R} \rightarrow \cbf{A, 1\text{ if }inv(R)\text{ else }2}$
    \STATE $\phi^{\text{ex}} := \bigwedge\limits_{\exists R.A \in \parts(\tbox)} \cbf{A, 1\text{ if }inv(R)\text{ else }2} \land \cbf{R} \rightarrow \cbf{\exists R.A, 2\text{ if }inv(R)\text{ else }1}$
    \STATE $\phi^{\text{inv}} := \bigwedge\limits_{R \in \roles} \cbf{R} \rightarrow \cbf{R^-}$
    \STATE $\cbf{\mathbb{D}_0} := \phi^{\text{kb}} \land \phi^{\text{uni}} \land \phi^{\text{ex}} \land \phi^{\text{inv}}$
    \RETURN $\cbf{\mathbb{D}_0}$
    \end{algorithmic}
\end{algorithm}

Algorithm \ref{alg:d0} constructs the boolean formula encoding the domino set $\mathbb{D}_0$. The predicate $inv(R)$ is true when $R$ is an inverse role \ie a role of the form $R^-$ and $\cbf{\cdot}$ is the characteristic boolean function defined as
\begin{equation}
    \cbf{X, i} = \begin{cases}
        \lnot \cbf{Y, i} &\text{ if } X = \lnot Y \\
        \cbf{Y, i} \land \cbf{Z, i} &\text{ if } X = Y \sqcap Z \\
        \cbf{Y, i} \lor \cbf{Z, i} &\text{ if } X = Y \sqcup Z \\
        \Vars[(X, i)]\text{ if }X \in \concepts\text{ else }\Vars[X] &\text{ otherwise} \\
    \end{cases} \label{eq:cbf}
\end{equation}

\begin{algorithm}
    \caption{Compute the circuit $\CKB = \cbf{\canonicaldomino}$ for the canonical domino set $\canonicaldomino$}
    \label{alg:d}
    \begin{algorithmic}
    \REQUIRE $\mathbb{D}_0$ the initial domino set
    \STATE i := 0
    \STATE $\mathscr{E}_1 := \mathbf{R} \cup \{ (X, 1) \mid X \in \parts(\ontology) \}$
    \STATE $\mathscr{E}_2 := \mathbf{R} \cup \{ (X, 2) \mid X \in \parts(\ontology) \}$
    \REPEAT
        \STATE $\phi_i^{\text{delex}} := \bigwedge\limits_{\exists R.U \in \parts(\ontology)} \bigwedge\limits_{j \in \{1, 2\}} \cbf{\exists R.U, j} \rightarrow \exists \mathscr{E}_j.(\cbf{\mathbb{D}_{i - 1}} \land \cbf{R} \land \cbf{C, 1\text{ if }inv(R)\text{ else }2})$
        
        \STATE $\phi_i^{\text{deluni}} := \bigwedge\limits_{\forall R.U \in \parts(\ontology)} \bigwedge\limits_{j \in \{ 1, 2 \}} \cbf{\forall R.U, j} \rightarrow \lnot \exists \mathscr{E}_j.(\cbf{\mathbb{D}_{i - 1}} \land \cbf{R} \land \cbf{C, 1\text{ if }inv(R)\text{ else }2})$
        
        \STATE $\cbf{\mathbb{D}_i} := \cbf{\mathbb{D}_{i - 1}} \land \phi_i^{\text{delex}} \land \phi_i^{\text{deluni}}$

        \STATE i := i + 1
    \UNTIL $\cbf{\mathbb{D}_i} = \cbf{\mathbb{D}_{i - 1}}$
    \RETURN $\cbf{\mathbb{D}_i}$
    \end{algorithmic}
\end{algorithm}

Finally, Algorithm \ref{alg:d} applies the iterative refinement on the canonical domino set produced by Algorithm \ref{alg:d0} by relying on existential quantification, which can be performed efficiently on SDDs \citep{darwiche2002knowledge,darwiche2011sdd}.

Note that, similarly to our approach, other methods \eg ProbLog \citep{de2007problog}, DeepProbLog \citep{manhaeve2018deepproblog}, Scallop \citep{li2023scallop}, BUNDLE \citep{riguzzi2013bundle}, rely on circuits obtained through knowledge compilation. Informally, they compile a boolean function that encodes a proof obtained using an automated reasoning software (e.g. a theorem prover or a deductive reasoner) and compile it to a circuit to tractably perform probabilistic reasoning \citep{chavira2008wmc}. Hence, the circuit produced encodes a \textit{local} representation of the knowledge of $\ontology$, which depends on the proof derived. The circuit $\Co$ obtained using Algorithm \ref{alg:d}, on the other hand, encodes a \textit{global} representation of the knowledge of $\ontology$, enabling applications such as sampling, as described in Section \ref{sec:intro}. We reserve a more formal analysis of the relationship between the circuit $\Co$ and other methods for future work.

\section{Experiments}
\label{sec:experiments}
In this section, we experiment with the different applications offered by $\Co$ as introduced in Section \ref{sec:dl-circuit}. In Section \ref{sec:data-generation}, we experiment with synthetic data generation using the circuit $\Co$ (\textbf{C1}). We show how $\Co$ can be harnessed to generate structured synthetic data that encapsulates the knowledge defined in $\ontology$. To do so, we define an algorithm that generates an ontology relying on the main modeling aspects of the $\ALCI$ fragment: concept disjointness and role restrictions. We then describe how one can construct a Knowledge Graph $\KG$ by sampling from $\Co$. Finally, we describe how to synthesize a dataset $\mathcal{D}$ based on the assertions of $\KG$. We qualitatively show that the distribution of $\mathcal{D}$ follows the knowledge in $\ontology$ and quantitatively evaluate the complexity of the synthesized data. In Section \ref{sec:logical-reasoning}, we experiment with $\Co$ as a deductive reasoner and compare its results with other search-based DL reasoners (\textbf{C2}). Empirically, we show that $\Co$ behaves asymptotically similarly to other reasoners. However, its parallelizable nature allows an efficient GPU-based implementation with a runtime up to three orders of magnitude faster on high volumes of data. Finally, in Section \ref{sec:nesy}, we experiment with NeSy models based on $\Co$ (\textbf{C3}). We generate a Knowledge Base $\KB$ using the method of Section \ref{sec:data-generation} and show that a NeSy approach is key to achieve reliable models consistent with the knowledge defined in the ontology.

All the experiments are executed on a Linux machine with an Intel i9-11900KF CPU with 128GB of RAM and an NVIDIA RTX 3090 with 24GB of RAM. All the code is available at \url{https://github.com/n28div/descriptionlogic}.

\subsection{Data generation}
\label{sec:data-generation}
Training and testing models using synthetic data is a widespread technique in Machine Learning \citep{patki2016sdv,jordon2022synth,dankar2022synth,hittmeir2019synth}, allowing researchers to focus on challenging aspects that are difficult to isolate in other datasets, such as reasoning shortcuts \citep{bortolotti2024rsbench} or spurious correlations \citep{qiu2024spuriouscomplexity}. Generating a challenging yet versatile synthetic dataset, however, is not a trivial task and requires a carefully engineered approach. In this section, we show that the circuit $\Co$ can be used to generate a dataset that encapsulates the domain knowledge encoded in $\ontology$. We first show how to generate an $\ALCI$ ontology (Section \ref{sec:ontology-generation}). We then show how to generate a synthetic Knowledge Graph $\KG$ based on the knowledge defined in the ontology $\O$ using $\Co$ (Section \ref{sec:kg-generation}). Finally, we describe how to synthesize a dataset from $\KG$ (Section \ref{sec:synthesize}), qualitatively show that it encapsulates the knowledge of $\ontology$, and quantitatively show the performances of different Machine Learning algorithms.

\subsubsection{Ontology generation}
\label{sec:ontology-generation}
Automatic generation of DL ontologies where specific properties, such as deeply nested hierarchical definitions, are enforced has been explored in literature as the result of some stochastic process where user-defined concepts and roles are automatically organized in an ontology \citep{elhaik1998generatedl,hubert2024pygraft,ma2006owlbench}.

\begin{algorithm}
    \caption{Randomly generate an $\ALCI$ ontology}
    \label{alg:generate-ontology}
    \begin{algorithmic}
    \REQUIRE Number of concepts $N_c$, number of roles $N_r$
    \REQUIRE Probability $p_d$, $p_r$, $p_c$ of domain and range restrictions and of concepts being disjoint

    \STATE $\concepts := \{ \mathsf{A_1}, \cdots, \mathsf{A_{N_c}} \}$
    \STATE $\roles := \{ \mathsf{R_1}, \cdots, \mathsf{R_{N_r}} \}$
    \STATE $\ontology := \concepts \cup \roles \cup \{ \mathsf{A} \sqsubseteq \lnot B \mid (A, B) \in \concepts \times \concepts, A \neq B, p \sim \mathcal{U}_{[0, 1]} < p_c \}$
    
    \FORALL{$\mathsf{R} \in \roles$}
        \STATE $\ontology := \ontology \cup \{ \mathsf{A} \sqsubseteq \mathsf{R}^-.(\sqcup_{\mathsf{C} \in \mathbf{X}} \mathsf{C}) \mid \mathbf{X} = \{ \mathsf{C} \mid \mathsf{C} \in \concepts, p \sim \mathcal{U}_{[0, 1]} < p_d \} \}$
        \STATE $\ontology := \ontology \cup \{ \mathsf{A} \sqsubseteq \mathsf{R}.(\sqcup_{\mathsf{C} \in \mathbf{X}} \mathsf{C}) \mid \mathbf{X} = \{ \mathsf{C} \mid \mathsf{C} \in \concepts, p \sim \mathcal{U}_{[0, 1]} < p_r \} \}$
    \ENDFOR
    \RETURN $\ontology$
    \end{algorithmic}
\end{algorithm}

In this work, we take inspiration from those methods and define Algorithm \ref{alg:generate-ontology}, which focuses on the main modeling features of the $\ALCI{}$ fragment: concept disjointness and role constraints \citep{rudolph2011dlfoundations}. The algorithm takes as input the number of roles $N_r$ and concepts $N_c$. Its generation process can be controlled through the parameters $p_d$, $p_r$, which are the probabilities that a concept appears, respectively, as a domain or range restriction of a role. Similarly, the parameter $p_c$ is used to control the probability that two concepts are disjoint. Algorithm \ref{alg:generate-ontology} does not produce any subsumption relation between different concepts. While this is also a key modeling aspect of all DL fragments, when coupled with concept disjunction, it can lead to inconsistent ontologies \ie, ontologies that do not admit any model. Other methods, such as the one proposed by \citet{hubert2024pygraft}, generate a possibly inconsistent ontology first and integrate deductive reasoners in the generation process to refine the ontology until consistency is obtained again. For the sake of simplicity, we maintain a simpler generation procedure, since it still allows us to experiment with the main requirements of our NeSy objective. Moreover, we remark that concept disjunction and role constraint are sufficient to model complex problems, such as the popular graph coloring \citep{gaggl2016fixeddomaindl}.

\begin{figure}
    \centering
    \includegraphics[width=1.0\linewidth]{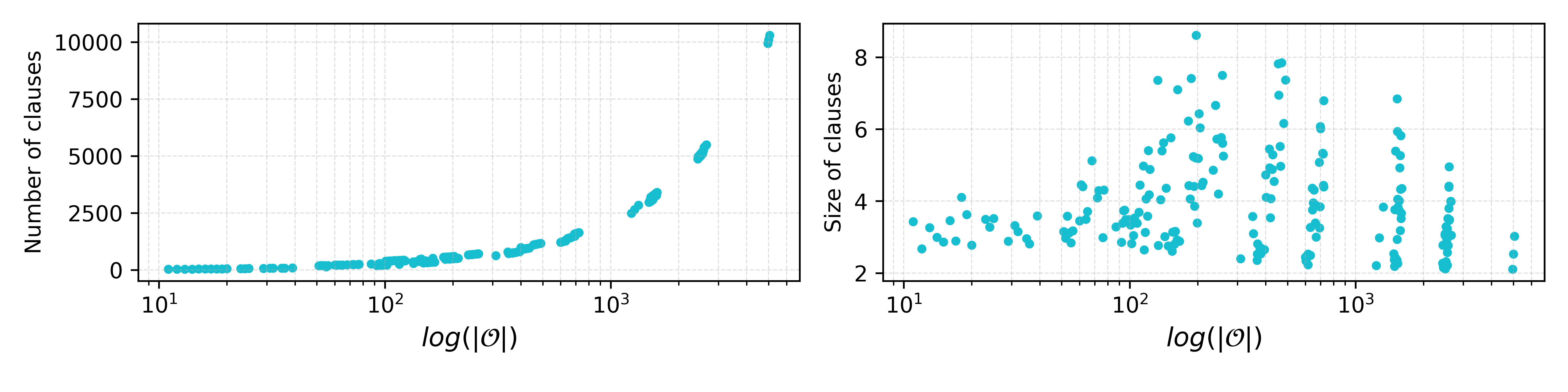}
    \caption{Dimension of the CNF formula generated by Algorithm \ref{alg:d0} for ontologies generated using Algorithm \ref{alg:generate-ontology} varying $N_c \in \{ 5, 10, 25, 50, 100 \}$, $N_r \in \{ 5, 25, 50 \}$, $p_d,p_r \in \{ 0.3, 0.5, 0.8 \}$, and $p_c \in \{ 0.3, 0.5, 1.0 \}$. The average size of clauses refers to the average number of atoms for each clause in the CNF.
    }
    \label{fig:ontology-cnf-dimensions}
\end{figure}

In principle, the method of Section \ref{sec:compilation} can compile any ontology generated using Algorithm \ref{alg:generate-ontology}. In practice, however, since the boolean formula produced by Algorithm \ref{alg:d} is exponentially big with respect to the number of parts of $\ontology$, large ontologies might produce large propositional formulas that are hard to compile. This is because compilation algorithms are time- and space-exponentially complex in the size of the formula \citep{darwiche2002knowledge}. In Figure \ref{fig:ontology-cnf-dimensions} we show how the number of clauses and variables for each clause grows in the CNF produced by Algorithm \ref{alg:d0} with respect to the size of $|\ontology|$ of the ontology. To do so, we generate a set of ontologies by varying the number of concepts $N_c \in \{ 5, 10, 25, 50, 100 \}$, the number of roles $N_r \in \{ 5, 25, 50 \}$ and all the combinations of the probabilities $p_d,p_r \in \{ 0.3, 0.5, 0.8 \}$ and $p_c \in \{ 0.3, 0.5, 1.0 \}$. While the average number of variables in a clause does not follow a clear behavior in function of the size of $\ontology$, the average number of clauses in the CNF is exponential in the dimension of the ontology. This aspect is important since it poses a concrete limit on the ontologies that can be compiled using our method.

\begin{figure}[htbp]
    \centering
    \includegraphics[width=\linewidth]{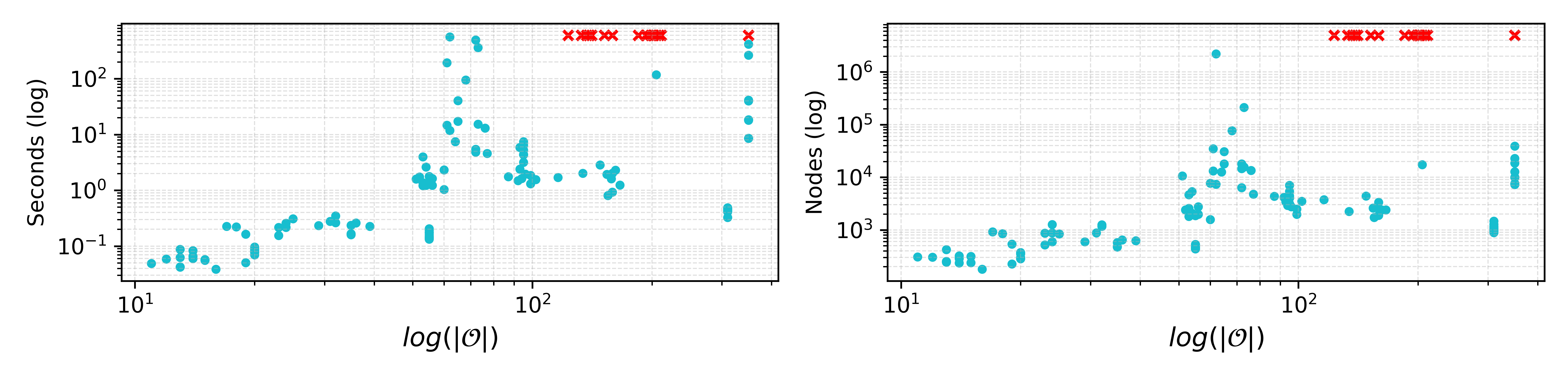}
    \caption{Time and number of nodes obtained by compiling $\ontology$, varying $N_c \in \{ 5, 10, 25 \}$ and $N_r \in \{ 5, 25 \}$ and averaged over all combinations of $p_d,p_r \in \{ 0.3, 0.5, 0.8 \}$, $p_c \in \{ 0.3, 0.5, 1.0 \}$. A timeout of $10$ minutes has been set on the compiler. Circuits that could not be compiled are shown as red crosses.}
    \label{fig:ontology-cnf-compiled}
\end{figure}

In Figure \ref{fig:ontology-cnf-compiled}, we show the time taken to compile the ontologies of Figure \ref{fig:ontology-cnf-dimensions} when relying on the original SDD knowledge compiler \citep{darwiche2011sdd}. For each formula, we impose a 10-minute timeout and measure the time taken to successfully compile the ontology and the number of nodes in the resulting circuit. Similarly to the results in Figure \ref{fig:ontology-cnf-dimensions}, the dimension of $\ontology$ and the time taken to compile it positively correlate with the size of the formula. Note, however, that it might be possible to rely on different knowledge compilers \citep{muise2012dsharp,oztok2015top-down-sdds,oztok2014cnfddnnf,lagniez2017d4,lai2025panini} to obtain better pragmatic performances. We reserve a more comprehensive analysis for future work.

\subsubsection{Knowledge Graph generation}
\label{sec:kg-generation}
In this section, we show that the circuit $\Co$ can be used to generate a Knowledge Graph $\KG$ based on the knowledge defined in the ontology $\ontology$.

Consider a sample $\mathbf{s}$ drawn from $\Co$ \ie $\mathbf{s} \sim \Co$, where the input units of $\Co$ are parameterized using indicator functions and the weights of sum units are initialized to one. Since the circuit encodes the canonical domino set $\canonicaldomino$, $\mathbf{s}$ corresponds to the encoding of some domino $\domino \in \canonicaldomino$ such that $\cbf{\domino} = \mathbf{s}$, where $\mathcal{A}$ and $\mathcal{B}$ refer to some individuals $a, b$ by definition of domino. Moreover, due to the characteristic boolean function $\cbf{\cdot}$ defined in Equation \eqref{eq:cbf}, the vector $\mathbf{s}$ can be seen as the concatenation operation $\mathbf{s} = [\mathbf{a}; \mathbf{r}; \mathbf{b}]$ where $\mathbf{a}$ (resp. $\mathbf{b}$) is a boolean vector representing the knowledge encoded in $\mathcal{A}$ about $a$ (resp. $\mathcal{B}$ and $b$) and $\mathbf{r}$ encodes the roles asserted between $a$ and $b$ as specified in $\mathcal{R}$.

In other words, sampling from $\Co$ means sampling two random individuals and some relationships that hold between them in some interpretation of $\ontology$. Clearly, the knowledge about both individuals (how they are classified) and their relations must be consistent with $\ontology$ by definition. 

By naively sampling from $\Co$ we assume that all the sampled individuals are different, due to the unique name assumption of DL. This produces a sparse $\KG$ where we only have pairwise relationships between entities. For example, in $\KG$ from Figure \ref{fig:ex:kb}, we would have that at most one artist is signed to each label. While this might be a desirable outcome in some settings, it defies the main purpose of Knowledge Graphs: to represent if and how different entities are related together \citep{hogan2022kgbook}.

To overcome this limitation, we first sample a set of individuals $\mathcal{E}$ from the circuit $\Co$ and then sample the relation holding between all their pairwise combinations.
The set of individuals $\mathcal{E}$ can be sampled from $\Co$ by sampling one boolean vector $\mathbf{e}_i$ for each $\varepsilon_i \in \mathcal{E}$, where $\mathbf{e}_i$ represents the knowledge encoded in some $\mathcal{A}$ of some domino $\domino \in \canonicaldomino$. 

Since we are only interested in the encoding of $\mathcal{A}$, we can avoid sampling the vectors related to $\mathcal{R}$ and $\mathcal{B}$. This corresponds to drawing samples from the circuit obtained by marginalizing the variables related to $\mathcal{R}$ and $\mathcal{B}$, which can be performed tractably in $\Co$ since it is decomposable \citep{choi2020pc,vergari2021compositional}.
Formally, we obtain $\mathbf{e}_i$ as 
\begin{equation*}
    \mathbf{e}_i \sim \sum_{R \in \roles} \sum_{C \in \concepts} \Co([\cbf{R}; \cbf{C, 2}])   
\end{equation*}
where the two sums over $\roles$ and $\concepts$ are effectively marginalizing some variables of $\Co$.

Similarly, we identify the relations between each combination of individuals $\mathbf{e}_1, \mathbf{e}_2 \in \mathcal{E}$ by using them as evidence when sampling from $\Co$. Sampling with evidence is also tractable on smooth and decomposable circuits such as SDDs \citep{choi2020pc}. The collection of all samples drawn using all combinations of individuals produces the final $\KG$. Formally, we have that $\KG$ is the set of vectors $\mathbf{a}_{ij}$ such that $\mathbf{a}_{ij} \sim \Co([\mathbf{e_i}; \mathbf{e_j}])$ for all $\mathbf{e}_i, \mathbf{e}_j \in \mathcal{E}$ with $i \leq j$. Notice that we also sample roles between an individual and itself in case $\ontology$ models reflexive relations.

\subsubsection{Synthesizing a dataset}
\label{sec:synthesize}
Given the generated $\ontology$ and $\KG$ as proposed before, we can construct a synthetic dataset $\dataset$ as initially described in Section \ref{sec:dl-circuit}. Informally, $\dataset$ contains \textit{observations} of the knowledge defined in a Knowledge Base $\KB$ \ie its samples are shaped as information that could be observed in the real world, such as images or text.

For instance, the sentence of example \ref{ex:sentence} is an observation of (some) of the knowledge of $\KB$ in Figure \ref{fig:ex:kb}. Observations contain latent information that is embedded in the semantics of their modality. For example, there is no direct mention of \textit{The Smiths} and \textit{Fugazi} being artists. This information is implied by common-sense knowledge instead. In this case, the ontology $\ontology$ acts as an explicit formalization of this latent common-sense knowledge. Clearly, an observation is a very general concept that spans any arbitrary modality, from images to embeddings produced by a pre-trained language model.

In this section, we propose a method that leverages the boolean vectors sampled from the circuit to synthesize the dataset of observations $\dataset$ from a $\KB$ generated in Section \ref{sec:data-generation}.

Recall that $\mathcal{E}$ refers to the individuals of $\KG$ and for each $\varepsilon_i \in \mathcal{E}$ we have a corresponding boolean vector representation $\mathbf{e}_i$ that we sampled from (marginalized) $\Co$.

We define the generative model $\mathcal{G}^\mathcal{E}_{i}$ which takes as input the vector $\mathbf{e}_i$ and outputs an observation of $\varepsilon_i$ in the form of an n-dimensional vector $\mathbf{o}_i \in \bbR^n$. In this work, we choose $\mathcal{G}^\mathcal{E}_i$ to be a multivariate normal distribution and use the vector $\mathbf{e}_i$ as its mean and a covariance $\Sigma_i$ \ie $\mathcal{G}^\mathcal{E}_i = \mathcal{N}(\mathbf{e}_i, \Sigma_i)$. In other words, for each individual, we define a normal distribution whose mean is its intensional description encoded as defined in Section \ref{sec:compilation}. Different individuals with the same intensional description will have different generative models through the use of different covariance matrices. It follows that each observation from the same individual will be generated from the same distribution, but will still display some superficial variability due to the covariance matrix. Moreover, samples that have a similar intensional description will have a more similar generating distribution. This allows observations with similar intensional descriptions to be represented with vectors that can be clustered together in principle.

Similarly, we define one generative model $\mathcal{G}^{\roles}$ for each possible combination of roles. In other words, we have a generative model $\mathcal{G}^{\roles}_K$ for each $K \in 2^{\roles}$. As done for individuals, we define $\mathcal{G}^{\roles}_K$ as a multivariate normal distribution, \ie $\mathcal{G}^{\roles}_K = \mathcal{N}(\mathbf{v}_K, \Sigma_K)$ where $\mathbf{v}_K$ is a $|\roles|$-dimensional boolean vector representing a multi-hot encoding of the combination $K$.

We can now synthesize the sample $\dataset^{(i)} \in \dataset$ as 
\begin{equation}
 \mathcal{D}^{(i)} = [ \mathbf{s} \sim \mathcal{G}^{\mathcal{E}}_i;\ \mathbf{r}_{ij} \sim \mathcal{G}^{\roles}_{ij};\ \mathbf{o} \sim \mathcal{G}^{\mathcal{E}}_j ]   
\end{equation}
for some individuals $\varepsilon_i, \varepsilon_j \in \mathcal{E}$ where $\mathbf{r}_{ij}$ is the multi-hot encoding of the relations between them. 

The dataset $\dataset$ is a collection of $n$-dimensional vectors with $n = |\concepts| + |\roles| + |\concepts|$.

\begin{figure}[htbp]
    \centering
    \begin{minipage}{0.4\textwidth}
        \centering
        \includegraphics[width=\linewidth]{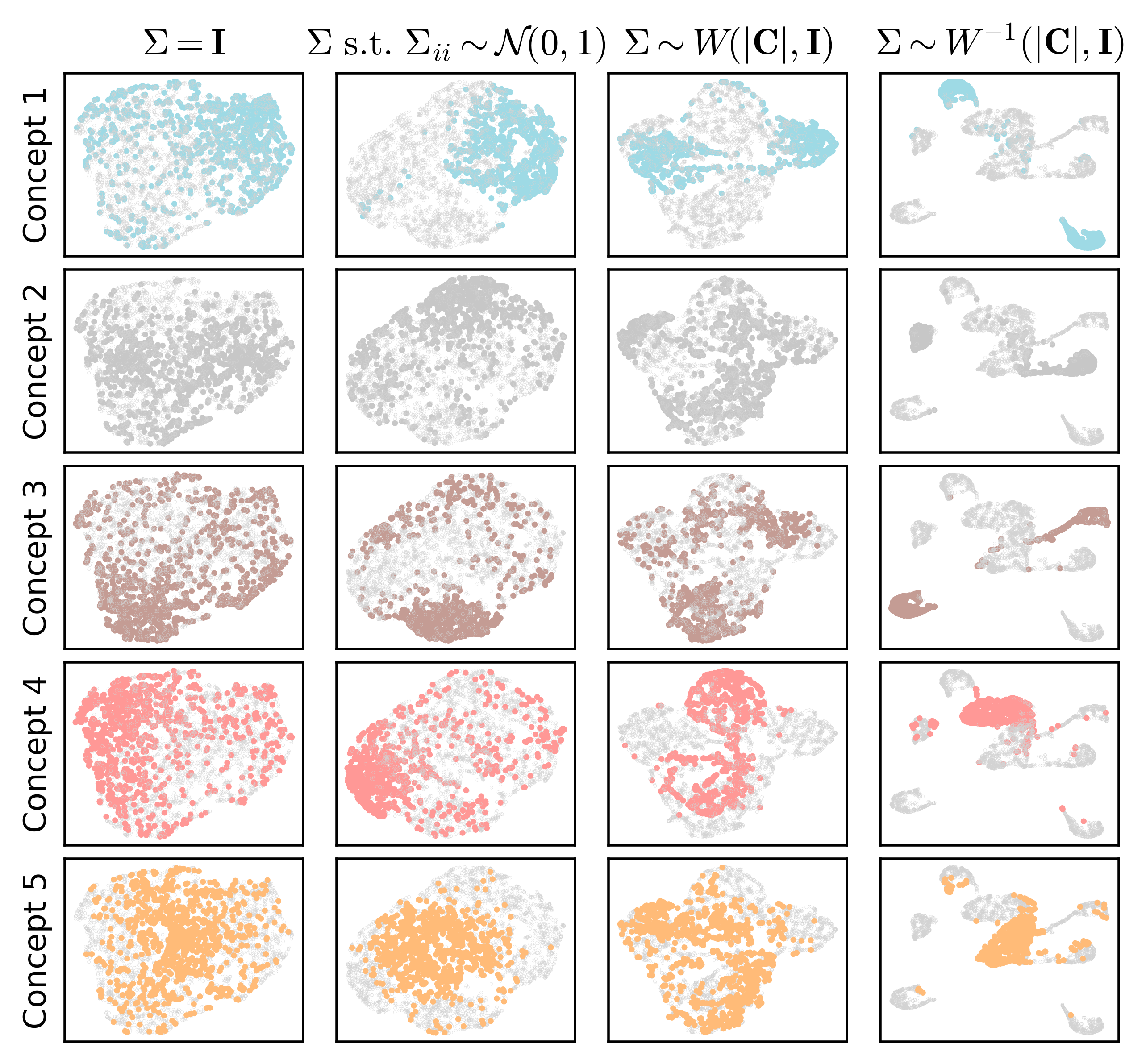}
        \caption{2D projection of $5k$ samples for an ontology generated with $N_c = 5$, $N_r = 3$, $p_d = p_r = 0.5$ and $p_c = 1$ using UMAP \citep{mcinnes2018umap}. $W$ and $W^{-1}$ refer to the Wishart distribution and its inverse.}
        \label{fig:gen:2d}
    \end{minipage}%
    \hfill
    \begin{minipage}{0.50\textwidth}
        \centering
        \begin{tabular}{lcccc}
        \toprule
         & \multicolumn{4}{c}{$\Sigma$} \\
        Model & $\mathbf{I}$ & $\mathcal{N}(0, 1)$ & $W$ & $W^{-1}$ \\
        \toprule
        Random & 0.20 & 0.20 & 0.20 & 0.20 \\
        DT & 0.33 & 0.63 & 0.56 & 0.89 \\
        NB & 0.50 & 0.75 & 0.42 & 0.73 \\
        kNN$_{10}$ & 0.44 & 0.71 & 0.67 & 0.93 \\
        kNN$_{100}$ & 0.49 & 0.70 & 0.65 & 0.89 \\
        RF & 0.45 & 0.72 & 0.68 & 0.93 \\
        LR & 0.50 & 0.64 & 0.35 & 0.53 \\
        SVM & 0.49 & 0.72 & 0.68 & 0.93 \\
        MLP & 0.40 & 0.67 & 0.64 & 0.93 \\
        \bottomrule
        \end{tabular}
        \captionof{table}{Classification results on the samples of Figure \ref{fig:gen:2d} using different Machine Learning methods where random predicts one random class with uniform probability, DT is a random tree, NB is Naive Bayes with Gaussian inputs, kNN$_n$ used the nearest $n$ neighbors for classification, RF is random forest, LR logistic regression, SVM is support vector machine with a rbf kernel, MLP has three hidden layers with 128 units and ReLU activations.}
        \label{tab:gen:classification}
    \end{minipage}
\end{figure}

Figure \ref{fig:gen:2d} shows how observations from the same individual are clustered together using different choices of $\Sigma$. To do so, we first generate an ontology $\ontology$ using Algorithm \ref{alg:generate-ontology} using $N_c = 5$, $N_r = 3$, $p_d = p_r = 0.5$ and $p_c = 1$. This produces a set of $5$ mutually disjoint concepts, since $p_c = 1$. We identify one single individual for each concept and sample $1k$ observations for each individual from its corresponding generative model. Finally, we use the UMAP \citep{mcinnes2018umap} dimensionality reduction method to plot the observations. We test four different choices of $\Sigma$: (i) the identity matrix ($\Sigma_i = \mathbf{I})$; (ii) the identity matrix where the main diagonal is replaced with values sampled from a normal distribution ($\Sigma_i$ such that $\Sigma_{jj} \sim \mathcal{N}(0, 1)$); (iii) matrices sampled from the Wishart ($\Sigma_i \sim Wishart(\mathbf{v}_i, \mathbf{I})$ with $\mathbf{v}_i$ the mean of the corresponding generative model); (iv) matrices sampled from the inverse Wishart distribution ($\Sigma_i \sim Wishart^{-1}(\mathbf{v}_i, \mathbf{I})$).

The figure illustrates how the generative model of each individual produces observations that are clustered together, where the density of clusters and their distance from other clusters are influenced by the covariance matrices. Relying on a $\Sigma = \mathbf{I}$ produces moderately clustered observations, while replacing its main diagonal with values sampled from a normal distribution produces clusters that are more neatly separated. Both choices, however, still partition the space into different distinguishable sections, making geometrical distance a reliable predictor of similarity between observations.

On the other hand, sampling the covariance matrix from a Wishart distribution produces observations where the distance is not a strong predictor of the similarity between observations, while sampling from the inverse Wishart distribution produces observations where clusters are highly separated.

All of those aspects influence the complexity of the dataset when used for training Machine Learning models. In Table \ref{tab:gen:classification}, we test different popular ML models and report their accuracy in predicting the observations' concept. For each individual, we train on $80\%$ of its observations and test on the remaining $20\%$. The classification task is a regular multiclass classification problem, since all concepts are mutually disjoint. Table \ref{tab:gen:classification} confirms the qualitative findings of Figure \ref{fig:gen:2d}. Observations generated using the inverse Wishart distribution, for example, are easier to classify when compared to observations generated using the identity matrix due to being more highly clustered.

\subsection{Deductive reasoning}
\label{sec:logical-reasoning}
The driving motivation of the different Description Logic fragments is to reduce the expressivity of First Order Logic languages so that decidable automated reasoning algorithms can be defined while maintaining a language expressive enough to model domain-specific knowledge \citep{baader2003dlhandbook,vanharmelen2008kr}. 
The time complexity of reasoning in most Description Logic fragments, however, is often exponential in the size of the ontology. This is a strong limit on the applicability of \textit{textbook} algorithms, particularly in real-world scenarios where one is interested in working with a large amount of data. In practice, this has been addressed by proposing search algorithms for which clever heuristics and optimizations can be used to obtain efficient implementations \citep{bail2010justbench}. A compilation-based approach aims at addressing this issue from a different angle. The exponential blow-up of the search space is encoded in a succinct representation using expensive knowledge compilation techniques. This is then amortized by the ability to execute different complex queries efficiently in the circuit.

Of course, the efficiency of compilation-based approaches is also dependent on the implementation of the circuit. The computational nature of circuits, however, allows the use of modern GPU-based parallel computations \citep{maene2025klay,peharz2020einsum,loconte2025circuitfactorization,liu2024pyjuice} that benefit from recent software and hardware advancements in the field. Informally, it is possible to evaluate multiple assignments at the same time by representing a state assignment as a vector and relying on parallelized implementations of sums and products on GPUs to evaluate multiple assignments at the same time.

\begin{figure}
    \centering
    \includegraphics[width=\linewidth]{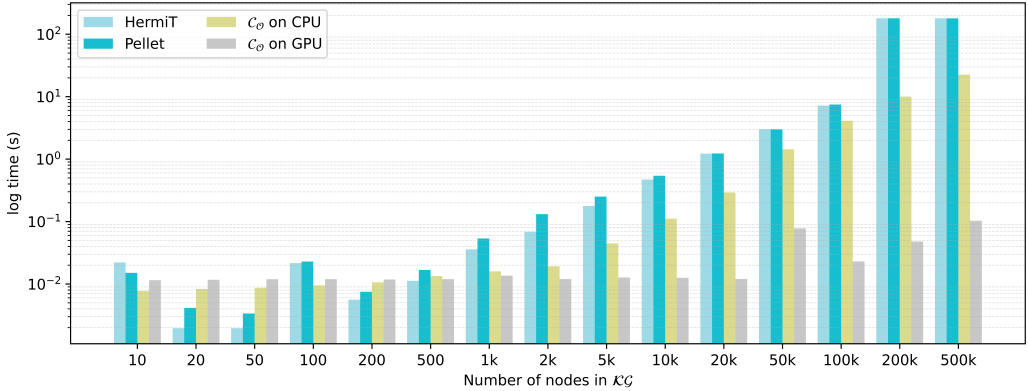}
    \caption{Deductive reasoning performances using CPU and GPU implementations of $\Co$ when compared to HermiT \citep{glimm2014hermit} and Pellet \citep{sirin2007pellet}. Each reasoner has been used with a timeout of $180$ seconds.}
    \label{fig:circuit-reasoning}
\end{figure}

In Figure \ref{fig:circuit-reasoning} we generate an ontology $\ontology$ using Algorithm \ref{alg:generate-ontology} with $N_c = 10$, $N_r = 5$ and $p_d = p_r = p_c = 0.5$. We then generate different $\KG$s varying the number of individuals as presented in Section \ref{sec:kg-generation} and check their consistency using CPU and GPU-based implementation of $\Co$ using \texttt{cirkit}\footnote{\url{https://github.com/april-tools/cirkit}}. We compare the runtime of $\Co$ with HermiT \citep{glimm2014hermit} and Pellet \citep{sirin2007pellet} by converting the generated KG to its corresponding RDF representation, as required by standard DL reasoners. For each KG, we set a runtime timeout of $180$ seconds. 

Note that Pellet and HermiT implement complex optimizations and address more complex DL fragments than $\ALCI$. Hence, the results should not be interpreted as a raw indication of the fastest reasoner, but rather to evaluate the scalability of compilation-based methods when compared to search-based methods. On moderately sized Knowledge Graphs containing less than 10k individuals, both search and compilation approaches obtain sub-second runtimes. The main difference, however, can be seen when the number of nodes increases. All methods exhibit a similar exponential blow-up in terms of time, as a consequence of the theoretical complexity. When $\Co$ is executed on a GPU, however, it displays a slower growing rate, resulting in a reasoning time that is almost three orders of magnitude faster when compared to the other methods. This shows the pragmatic convenience of vectorized circuits executed on a GPU and paves the way to the development of deductive reasoners that directly exploit this feature.

\subsection{Reliable NeSy classification}
\label{sec:nesy}
In this Section, we show that the circuit $\Co$ can be used for effective and reliable multilabel NeSy classification by experimenting with Semantic Losses (SL) \citep{xu2018semantic} and Semantic Probabilistic Layers (SPL) \citep{ahmed2022semantic}.
Both approaches compile user-defined constraints as circuits using knowledge compilation techniques and integrate them within a neural network, leveraging the differentiability of circuits. We will first describe those methods and later report on the results of our experiments.

\paragraph{SL and SPL background}
To describe both the SL and SPL approaches, we consider a multilabel classification problem where we have an input $\mathbf{x}$ that has been labeled by an $L$-dimensional boolean vector $\yhat \in \{ 0, 1 \}^L$, with $L$ the number of classes. 

Similarly to \citet{ahmed2022semantic}, we observe that any neural network classifier can be abstracted as the combination of a feature extraction function $\nn$ that produces an $n$-dimensional feature vector $\mathbf{z} = \nn(\mathbf{x})$ from $\mathbf{x}$ and a classification function that predicts the probability distribution $p(\mathbf{y} | \mathbf{z})$. 

In a general setting, $\nn$ is trained to optimize a discriminative loss, such as minimizing the cross-entropy loss $\mathcal{L}_{CE}(\yhat, \mathbf{z)}$, where the different $L$ classes are assumed to be independent. In practice, we have that $p(\mathbf{y} | \mathbf{z})$ is implemented as
\begin{equation}
    p(\mathbf{y} | \mathbf{z}) = \prod_{i=0}^L p(\mathbf{y}_i = 1 | \mathbf{z}) \quad\quad p(\mathbf{y}_i = 1 | \mathbf{z}) = \sigma(\mathbf{w}^T \mathbf{z)}
    \label{eq:fil}
\end{equation}
with $\mathbf{w} \in \mathbb{R}^L$ a parameter vector of the classification function and $\sigma(x) = 1 / (1 + e^{-x})$ the logistic sigmoid activation function. 

The SL approach integrates the user constraints by updating the loss function $\mathcal{L}$ to 
\begin{equation}
    \mathcal{L} = \mathcal{L}_{CE}(\yhat, \mathbf{z)} + \lambda \mathcal{L}_{SL}(\mathbf{z}) \quad\quad \mathcal{L}_{SL}(\mathbf{z}) = \sum_{\mathbf{y}^* \in \{0, 1\}^L} p(\mathbf{y}^*|\mathbf{z}) \Co(\mathbf{y}^*)
    \label{eq:sl}
\end{equation}
where $\mathcal{L}_{SL}(\mathbf{z})$ is the Semantic Loss term and $\lambda$ is a parameter used to regulate its importance.
The SL terms relies on the circuit $\Co$ parameterized as done in Section \ref{sec:reasoning} (indicator function for input units and unitary weights for sum units) and computes the Weighted Model Count (WMC) with respect to the models of $\ontology$, which is tractable for smooth, decomposable and deterministic circuits \citep{chavira2008wmc}.

In a nutshell, the SL enumerates all the consistent assignments of $\Co$ and weights them according to the belief predicted by the neural network. Since the support $\csupport{\Co}$ matches the models of $\ontology$, the probability mass that $\nn$ places on inconsistent assignments is discarded. By maximizing $\mathcal{L}_{SL}$, one seeks to instruct $\nn$ to place more probability mass on consistent assignments. This interpretation suggests that higher values of $\lambda$ lead to an optimization procedure that focuses more on predicting consistent assignments, which might disadvantage the optimization of $\mathcal{L}_{CE}$. The SL is hence treated as a regularization term, where the parameter $\lambda$ is a hyper-parameter to optimize based on the downstream performances \citep{xu2018semantic}. Note that this can be useful in settings where the knowledge represented in $\ontology$ is noisy and therefore one wishes to enforce \textit{soft-constraints} on the model, by relying on small values of $\lambda$.

Different from SL, a Semantic Probabilistic Layer \citep{ahmed2022semantic} directly relies on the circuit $\Co$ to implement the probability distribution $p(\mathbf{y} | \mathbf{z})$. Intuitively, since the support of $\Co$ encodes the set of dominoes that are consistent with $\ontology$ by definition, \citet{ahmed2022semantic} proposes to optimize $\nn$ so that it outputs a parameterization of $\Co$ such that its most probable state is $\yhat$. 

Formally, an SPL computes $p(\mathbf{y} | \mathbf{z})$ as
\begin{equation}
    p(\mathbf{y} | \mathbf{z}) = \Co(\mathbf{y}; \mathbf{\Omega} = g(\mathbf{z}))
    \label{eq:spl}
\end{equation}
where $\mathbf{\Omega}$ is the set of parameters of the sum units of $\Co$ and $g$ is called a gating function \citep{shao2022conditionalgatefunction}, which takes as input the feature vector $\mathbf{z}$ and outputs a set of parameters $\mathbf{\Omega}$. The gating function can be implemented using any arbitrary function, including the identity function or training a specific neural network model. Note that for $\Co(\mathbf{y}; \mathbf{\Omega} = g(\mathbf{z}))$ to represent a valid probability distribution, the parameters of each sum unit must be non-negative and sum up to one \citep{peharz2015theoretical}, which can be done by relying on a softmax activation for each sum units.

Rather than optimizing a discriminative loss, one minimizes the negative log-likelihood
\begin{equation}
    \mathcal{L} = - \log(p(\yhat | \mathbf{z})) = - \log(\Co(\yhat; \mathbf{\Omega} = g(\mathbf{z})))
\end{equation}
Informally, one trains $\nn$ to output parameters for $\Co$ that maximize the belief over the assignment $\yhat$ when $\mathbf{z}$ is observed.

At inference time, it is possible to output predictions by extracting the most probable assignment $\mathbf{y}$ \ie query $\Co$ for its MAP (Maximum A Posteriori) assignment, as
\begin{equation}
    \mathbf{y} = \argmax\limits_{\mathbf{y}^* \in \{0, 1\}^L} \Co(\mathbf{y}^*; \mathbf{\Omega} = g(\mathbf{z}))
    \label{eq:spl-map}
\end{equation}
which is tractable for smooth, decomposable, and deterministic circuits \citep{choi2020pc,darwiche2009modeling}.
Different from the SL approach, an SPL is guaranteed to produce a prediction $\mathbf{y}$ that is consistent with the ontology $\ontology$ since $\mathbf{y} \in \csupport{\Co}$ by definition and can only implement \textit{hard-constraints}.

Note that given their definitions in Equations \ref{eq:sl} and \ref{eq:spl}, both SL and SPL output predictions $\mathbf{y}$ that can be interpreted as vectors encoded using the characteristic boolean function of Equation \ref{eq:cbf} and can be interpreted accordingly, \ie each element of the vector refers to the and element of a domino.

Alongside the differences highlighted so far, SL and SPL also differ in their expressiveness. Informally, the independence assumption required by the SL approach results in models that are harder to train and produce over-confident predictions only over a small set of possible consistent assignments, possibly suffering from reasoning shortcuts \citep{krieken2024independence,marconato2024not}. On the other hand, an SPL models an expressive probability over the $L$ labels and is hence able to capture conditional dependencies between them.

\paragraph{Experiments using $\ontology$ and $\dataset$}
In this section, we experiment by generating the ontology $\ontology$ and the dataset $\dataset$ using the method of Section \ref{sec:data-generation} and setting $N_c = 10$, $N_r =5$, $p_d = p_r = 0.5$ and $p_c = 1.0$. Unlike the experiments reported in Table \ref{tab:gen:classification}, we frame the problem as a multilabel classification task where one must jointly classify the subject and object individuals and identify the roles that hold between them. 

For example, if we assume $\mathbf{x}$ encodes a representation of the sentence of Example \ref{ex:sentence} then the objective is to identify that the subject individual (\textit{Fugazi}) and the object individual (\textit{The Smiths}) are classified as $\artistConcept$, while only $\influeceRole$ holds between them.

We compare the SL and SPL approaches with an MLP baseline that predicts the subject, roles, and object of an input observation, akin to the MLP of Table \ref{tab:gen:classification}. Moreover, we also compare our approach with DeepProbLog \citep{manhaeve2018deepproblog}. 
Note that, as explained in Section \ref{sec:related}, reducing DL semantics to Prolog (and hence DeepProbLog) is not trivial.

\lstdefinestyle{listingstyle}{
    commentstyle=\color{olive},
    keywordstyle=\color{teal},
    numberstyle=\tiny\color{gray},
    basicstyle=\ttfamily\footnotesize,
    captionpos=b,                    
    keepspaces=true,                 
    numbers=left,
    xleftmargin=.1\textwidth, 
    xrightmargin=.1\textwidth
}
\lstset{
    style=listingstyle, 
    language=Prolog,
    morekeywords={nn},
}
\begin{lstlisting}[caption={Example of DeepProbLog formalization where only one role (r0) and two disjoint classes (c0, c1) are defined. The class c0 is domain of the role and the class c1 is range of the role.},label=code:dpl]
nn(nn_r0, [X]) :: r0(X) . % detect relation_0 in the input
% classify the subject 
nn(nn_c0_s, [X]) :: c0(subject(X)) . 
nn(nn_c1_s, [X]) :: c1(subject(X)) .
% classify the object
nn(nn_c0_o, [X]) :: c0(object(X)) .
nn(nn_c1_o, [X]) :: c1(object(X)) . 
not(c1(X)) :- c0(X) . % class 0 and class 1 are mutually disjoint
not(r0(X)) :- not(c0(subject(X))) . % the domain of r0 is C0
not(r0(X)) :- not(c1(object(X))) . % the range of r0 is C1
\end{lstlisting}

For this reason, we model the DeepProbLog logic program directly after the characteristics of the generated $\ontology$, mostly relying on the fact that setting $p_c = 1.0$ implies that all concepts in the ontology are mutually disjoint. A simplified version of our formalization using two concepts and one role is reported in Listing \ref{code:dpl}.

To obtain a similar expressivity between the baseline and the NeSy approaches, we design $\nn$ with two hidden layers with $128$ neurons and ReLU activations. We train all models for $30$ epochs using Adam \citep{kingma2014adam} and a fixed learning rate of $0.001$.

\begin{table}[ht]
    \centering
    \tiny
    \begin{tabular}{llllllll}
    \toprule
    $\Sigma \sim $ & Model & $\lambda$ & Precision $\uparrow$ & Recall $\uparrow$ & F1 $\uparrow$ & Exact Match $\uparrow$ & Consistent $\uparrow$ \\
    \midrule

    \multirow[c]{6}{*}{$\mathbf{I}$} & MLP & & $0.691 \pm 0.003$ & $0.617 \pm 0.006$ & $0.651 \pm 0.002$ & $0.134 \pm 0.005$ & $0.291 \pm 0.002$ \\
    & DPL \cite{manhaeve2018deepproblog} & & $0.479 \pm 0.021$ & $0.457 \pm 0.061$ & $0.444 \pm 0.033$ & $0.030 \pm 0.021$ & $\mathbf{1.000} \pm 0.000$ \\
    & \multirow[c]{3}{*}{SL \cite{xu2018semantic}} & 0.01 & $\mathbf{0.700} \pm 0.005$ & $0.601 \pm 0.003$ & $0.645 \pm 0.001$ & $0.132 \pm 0.007$ & $0.993 \pm 0.003$ \\
    &  & 0.001 & $0.695 \pm 0.003$ & $0.608 \pm 0.003$ & $0.647 \pm 0.001$ & $0.135 \pm 0.007$ & $0.978 \pm 0.004$ \\
    & & 0.0001 & $0.696 \pm 0.003$ & $0.609 \pm 0.001$ & $0.648 \pm 0.001$ & $0.136 \pm 0.008$ & $0.981 \pm 0.003$ \\
    & SPL \cite{ahmed2022semantic} & & $0.667 \pm 0.005$ & $\mathbf{0.666} \pm 0.003$ & $\mathbf{0.666} \pm 0.004$ & $\mathbf{0.219} \pm 0.008$ & $\mathbf{1.000} \pm 0.000$ \\
    \midrule

    \multirow[c]{6}{*}{$\mathcal{N}(0, 1)$} & MLP & & $0.735 \pm 0.006$ & $0.646 \pm 0.005$ & $0.687 \pm 0.000$ & $0.306 \pm 0.006$ & $0.307 \pm 0.008$ \\
    & DPL \cite{manhaeve2018deepproblog} & & $0.359 \pm 0.023$ & $0.310 \pm 0.133$ & $0.294 \pm 0.066$ & $0.051 \pm 0.003$ & $0.998 \pm 0.000$ \\
    & \multirow[c]{3}{*}{SL \cite{xu2018semantic}} & 0.01 & $0.741 \pm 0.008$ & $0.640 \pm 0.012$ & $0.685 \pm 0.004$ & $0.306 \pm 0.006$ & $0.985 \pm 0.006$ \\
    &  & 0.001 & $0.737 \pm 0.005$ & $0.645 \pm 0.011$ & $0.687 \pm 0.005$ & $0.309 \pm 0.005$ & $0.979 \pm 0.005$ \\
    & & 0.0001 & $0.736 \pm 0.003$ & $0.648 \pm 0.013$ & $0.688 \pm 0.006$ & $0.311 \pm 0.011$ & $0.979 \pm 0.006$ \\
    & SPL \cite{ahmed2022semantic} & & $\mathbf{0.783} \pm 0.007$ & $\mathbf{0.767} \pm 0.006$ & $\mathbf{0.775} \pm 0.005$ & $\mathbf{0.500} \pm 0.007$ & $\mathbf{1.000} \pm 0.000$ \\ 
    \midrule
    
    \multirow[c]{6}{*}{$W$} & MLP & & $0.440 \pm 0.012$ & $0.250 \pm 0.007$ & $0.280 \pm 0.005$ & $0.061 \pm 0.005$ & $0.007 \pm 0.003$ \\
    & DPL \cite{manhaeve2018deepproblog} & & $0.242 \pm 0.017$ & $0.272 \pm 0.119$ & $0.216 \pm 0.052$ & $0.021 \pm 0.003$ & $\mathbf{1.000} \pm 0.000$ \\
    & \multirow[c]{3}{*}{SL \cite{xu2018semantic}} & 0.01 & $0.461 \pm 0.001$ & $0.234 \pm 0.014$ & $0.267 \pm 0.006$ & $0.063 \pm 0.004$ & $\mathbf{1.000} \pm 0.000$ \\
    &  & 0.001 & $0.446 \pm 0.007$ & $0.243 \pm 0.010$ & $0.276 \pm 0.002$ & $0.062 \pm 0.006$ & $0.998 \pm 0.002$ \\
    & & 0.0001 & $0.444 \pm 0.004$ & $0.243 \pm 0.010$ & $0.278 \pm 0.004$ & $0.061 \pm 0.003$ & $0.999 \pm 0.001$ \\
    & SPL \cite{ahmed2022semantic} & & $\mathbf{0.541} \pm 0.006$ & $\mathbf{0.520} \pm 0.012$ & $\mathbf{0.530} \pm 0.009$ & $\mathbf{0.256} \pm 0.005$ & $\mathbf{1.000} \pm 0.000$ \\ \midrule
    
    \multirow[c]{6}{*}{$W^{-1}$} & MLP & & $0.666 \pm 0.005$ & $0.455 \pm 0.016$ & $0.536 \pm 0.010$ & $0.164 \pm 0.004$ & $0.133 \pm 0.015$ \\
    & DPL \cite{manhaeve2018deepproblog} & & $0.396 \pm 0.024$ & $0.363 \pm 0.163$ & $0.351 \pm 0.085$ & $0.021 \pm 0.007$ & $0.998 \pm 0.000$ \\
    & \multirow[c]{3}{*}{SL \cite{xu2018semantic}} & 0.01 & $\mathbf{0.675} \pm 0.010$ & $0.432 \pm 0.011$ & $0.521 \pm 0.004$ & $0.148 \pm 0.003$ & $0.998 \pm 0.002$ \\
    &  & 0.001 & $0.667 \pm 0.009$ & $0.447 \pm 0.014$ & $0.531 \pm 0.006$ & $0.152 \pm 0.008$ & $0.992 \pm 0.004$ \\
    & & 0.0001 & $0.664 \pm 0.007$ & $0.449 \pm 0.013$ & $0.531 \pm 0.005$ & $0.151 \pm 0.010$ & $0.993 \pm 0.005$ \\
    & SPL \cite{ahmed2022semantic} & & $0.645 \pm 0.011$ & $\mathbf{0.605} \pm 0.002$ & $\mathbf{0.624} \pm 0.006$ & $\mathbf{0.291} \pm 0.009$ & $\mathbf{1.000} \pm 0.000$ \\
\bottomrule
\end{tabular}
    \vspace{1em}
    \caption{Results on the joint multilabel classification task averaged over three different runs. NeSy models consistently outperform the neural baseline on the number of consistent predictions, while maintaining comparable performances. In particular, the SPL approach largely outperforms the other methods in almost all datasets and metrics, while guaranteeing logically consistent predictions.}
    \label{tab:nesy:results}
\end{table}

Table \ref{tab:nesy:results} summarizes the results obtained by averaging each metric over three random initializations. We generate four different datasets containing $100$ individuals and use the covariance matrices presented in Section \ref{sec:data-generation}. We train on $1k$ samples and report micro-averaged precision, recall, and F1 scores on a test set of $500$ held-out samples. We also report the exact match and ratio of consistent predictions for each model and dataset combination to evaluate the global performance and reliability of each method.

Results clearly show that NeSy models are more reliable when compared to the unconstrained neural baseline and produce consistent predictions regardless of the complexity of the dataset. As explained before, the SPL approach provably outputs consistent predictions, making it the most reliable NeSy model. Nonetheless, the Semantic Loss and DeepProblog approaches also produce mostly consistent predictions ($> 97\%$ in all datasets). In particular, it is possible to appreciate that higher values of $\lambda$ in SL-based models negatively correlate with the exact match performances \ie the discriminative power of the model is hampered by the SL regularization, but positively correlate with the consistency of the predictions, delineating a clear trade-off between downstream performances and reliability when a \textit{soft-constraint} approach is enforced.

On the other hand, in the best-case scenario, the neural baseline only produces consistent predictions in roughly one out of three samples. We highlight that on the dataset generated by sampling $\Sigma$ from a Wishart distribution, the results produced by the baseline are completely unreliable, which is also reflected in the quantitative metrics.

Overall, the results of Table \ref{tab:nesy:results} confirm the hypothesis of \citet{marcus2020robustai} illustrated in Section \ref{sec:intro}: NeSy approaches are an effective way of obtaining reliable deep learning models, since they exploit human knowledge while maintaining the required expressivity to model complex features.

\paragraph{Integrate background knowledge}
The experiments described above jointly perform the classification of the subject and object of the input and detect the relationships between them. While this approach is consistent with recent advancements in automated information extraction methods \citep{orlando2024relik,lou2023uie}, there are situations in which one has some background knowledge at hand that can be used, as also argued in Section \ref{sec:dl-circuit}. In this section, we demonstrate that background knowledge can be easily integrated into models based on SL or SPL. In particular, we update our experimental setting and assume that the classification of subjects and objects can be performed using background knowledge, while the detection of the relations is performed by the neural network, similarly to the link prediction setting described in Section \ref{sec:dl-circuit}.

We first consider the SL approach by recalling that $p(\mathbf{y} | \mathbf{z})$ is implemented as formalized in Equation \ref{eq:fil}.
We can inject background knowledge in $p(\mathbf{y} | \mathbf{z})$ by updating it to
\begin{equation}
    p(\mathbf{y} | \mathbf{z}) = \prod_{i \in \mathbf{L}_S} \llbracket \mathbf{e}^S_i = 1 \rrbracket \prod_{i \in \mathbf{L}_R} p(\mathbf{y}_i = 1 | \mathbf{z}) \prod_{i \in \mathbf{L}_O} \llbracket \mathbf{e}^O_i = 1 \rrbracket
    \label{eq:fil-bg}
\end{equation}
where $\mathbf{L}_S, \mathbf{L}_R, \mathbf{L}_O$ are, respectively, the indices that refer to subject, role, and object variables and $\mathbf{e}^S \in \{0, 1 \}^{|\mathbf{L}_S|}, \mathbf{e}^O \in \{0, 1 \}^{|\mathbf{L}_O|}$ are the evidence vectors encoding background knowledge according to the characteristic boolean function of Equation \ref{eq:cbf}. Intuitively, one constructs the classification function such that it contributes only to a subset of the probability distribution $p(\mathbf{y} | \mathbf{z})$, while background knowledge is used for the rest.

Integrating background knowledge in the SPL approach is even simpler. It is sufficient to update Equation \ref{eq:spl-map} to
\begin{equation}
    \mathbf{y} = \argmax\limits_{\mathbf{y}^* \in \{0, 1\}^L} \Co(\mathbf{y}^* | \mathbf{e}; \mathbf{\Omega} = g(\mathbf{z}))
    \label{eq:spl-map-bg}
\end{equation}
where $\mathbf{e}$ refers to the evidence vector encoding background knowledge according to the characteristic boolean function of Equation \ref{eq:cbf}. This is possible since querying for the MAP state using evidence is tractable in smooth, decomposable, and deterministic circuits \citep{choi2020pc,darwiche2009modeling}.

\begin{table}[ht]
    \centering
    \tiny
    \begin{tabular}{llllllll}
    \toprule
    $\Sigma \sim $ & Model & $\lambda$ & Precision $\uparrow$ & Recall $\uparrow$ & F1 $\uparrow$ & Exact Match $\uparrow$ & Consistent $\uparrow$ \\
    \midrule

    \multirow[c]{5}{*}{$\mathbf{I}$} & MLP & & $0.665 \pm 0.006$ & $0.680 \pm 0.022$ & $0.672 \pm 0.008$ & $0.145 \pm 0.005$ & $0.771 \pm 0.017$ \\
     & \multirow[c]{3}{*}{SL \cite{xu2018semantic}} & 0.01 & $0.666 \pm 0.006$ & $0.678 \pm 0.020$ & $0.672 \pm 0.007$ & $0.145 \pm 0.005$ & $0.767 \pm 0.013$ \\
     &  & 0.001 & $0.666 \pm 0.006$ & $0.680 \pm 0.020$ & $0.673 \pm 0.007$ & $0.146 \pm 0.003$ & $0.770 \pm 0.017$ \\
     & & 0.0001 & $0.665 \pm 0.004$ & $0.678 \pm 0.023$ & $0.671 \pm 0.010$ & $0.143 \pm 0.008$ & $0.767 \pm 0.013$ \\
     & SPL \cite{ahmed2022semantic} & & $\mathbf{0.716} \pm 0.005$ & $\mathbf{0.760} \pm 0.005$ & $\mathbf{0.737} \pm 0.001$ & $\mathbf{0.231} \pm 0.014$ & $\mathbf{1.000} \pm 0.000$ \\
     \midrule
    
    \multirow[c]{5}{*}{$\mathcal{N}(0, 1)$} & MLP & & $0.733 \pm 0.017$ & $0.742 \pm 0.019$ & $0.737 \pm 0.007$ & $0.278 \pm 0.014$ & $0.788 \pm 0.019$ \\
     & \multirow[c]{3}{*}{SL \cite{xu2018semantic}} & 0.01 & $0.736 \pm 0.019$ & $0.739 \pm 0.016$ & $0.737 \pm 0.009$ & $0.279 \pm 0.016$ & $0.791 \pm 0.022$ \\
     &  & 0.001 & $0.734 \pm 0.018$ & $0.740 \pm 0.018$ & $0.737 \pm 0.007$ & $0.277 \pm 0.014$ & $0.788 \pm 0.025$ \\
     & & 0.0001 & $0.735 \pm 0.016$ & $0.742 \pm 0.016$ & $0.738 \pm 0.006$ & $0.278 \pm 0.009$ & $0.790 \pm 0.020$ \\
     & SPL \cite{ahmed2022semantic} & & $\mathbf{0.767} \pm 0.004$ & $\mathbf{0.776} \pm 0.006$ & $\mathbf{0.772} \pm 0.005$ & $\mathbf{0.336} \pm 0.009$ & $\mathbf{1.000} \pm 0.000$ \\
     \midrule

    \multirow[c]{5}{*}{$W$} & MLP & & $0.630 \pm 0.002$ & $0.607 \pm 0.013$ & $0.618 \pm 0.006$ & $0.108 \pm 0.003$ & $0.687 \pm 0.003$ \\
     & \multirow[c]{3}{*}{SL \cite{xu2018semantic}} & 0.01 & $0.634 \pm 0.001$ & $0.601 \pm 0.014$ & $0.617 \pm 0.007$ & $0.113 \pm 0.008$ & $0.693 \pm 0.010$ \\
     &  & 0.001 & $0.633 \pm 0.004$ & $0.608 \pm 0.017$ & $0.620 \pm 0.011$ & $0.109 \pm 0.009$ & $0.691 \pm 0.003$ \\
     & & 0.0001 & $0.631 \pm 0.001$ & $0.607 \pm 0.013$ & $0.619 \pm 0.007$ & $0.109 \pm 0.008$ & $0.682 \pm 0.004$ \\
     & SPL \cite{ahmed2022semantic} & & $\mathbf{0.713} \pm 0.010$ & $\mathbf{0.677} \pm 0.045$ & $\mathbf{0.694} \pm 0.023$ & $\mathbf{0.177} \pm 0.032$ & $\mathbf{1.000} \pm 0.000$ \\
     \midrule

    \multirow[c]{5}{*}{$W^{-1}$} & MLP & & $\mathbf{0.822} \pm 0.013$ & $0.748 \pm 0.012$ & $0.783 \pm 0.001$ & $0.469 \pm 0.004$ & $0.889 \pm 0.017$ \\
     & \multirow[c]{3}{*}{SL \cite{xu2018semantic}} & 0.01 & $0.819 \pm 0.010$ & $0.743 \pm 0.016$ & $0.779 \pm 0.005$ & $0.464 \pm 0.005$ & $0.889 \pm 0.012$ \\
     &  & 0.001 & $0.818 \pm 0.010$ & $0.743 \pm 0.012$ & $0.779 \pm 0.003$ & $0.465 \pm 0.008$ & $0.885 \pm 0.014$ \\
     &  & 0.0001 & $0.820 \pm 0.012$ & $0.749 \pm 0.016$ & $0.783 \pm 0.003$ & $\mathbf{0.472} \pm 0.000$ & $0.890 \pm 0.014$ \\
     & SPL \cite{ahmed2022semantic} & & $0.808 \pm 0.005$ & $\mathbf{0.796} \pm 0.002$ & $\mathbf{0.802} \pm 0.004$ & $0.462 \pm 0.011$ & $\mathbf{1.000} \pm 0.000$ \\
\bottomrule
\end{tabular}
    \vspace{1em}
    \caption{Results on the link prediction task averaged over three different runs. NeSy models consistently outperform the neural baseline on the number of consistent predictions on the link prediction task while maintaining comparable performance with the neural baseline. Similarly to Table \ref{tab:nesy:results}, the SPL approach largely outperforms the other methods in almost all datasets and metrics, while guaranteeing logically consistent predictions.}
    \label{tab:nesy:results-bg}
\end{table}

Table \ref{tab:nesy:results-bg} shows the results obtained using the same experimental setting as the previous experiments. Different from the results of Table \ref{tab:nesy:results}, only the SPL approach consistently outperforms the MLP baseline on the number of consistent predictions, whereas the SL approaches do not result in a definite improvement over the baseline. Given the definition of the probability distribution in Equation \ref{eq:fil-bg}, the SL computation is biased towards consistent predictions; hence, an incorrect distribution of probability mass over inconsistent predictions is less influential than compared to the joint prediction task. On the other hand, SPL always achieves consistent predictions and, since we only update the inference phase, the model is still able to capture complex relations between subjects, objects, and roles during training. This allows it to outperform the other models in almost all of our experiments.

\section{Conclusion}
\label{sec:conclusion}
In this paper, we propose to compile an $\ALCI{}$ Description Logic ontology into a circuit representation to obtain deep learning models that can reliably output predictions consistent with the domain knowledge formalized in the ontology. We experiment with different NeSy methods and show that they obtain reliable performances and even outperform fully neural baselines, including cases when background knowledge is available.

Moreover, we show that challenging tasks can be addressed efficiently as a byproduct of compiling the ontology to a circuit, such as generating semantically defined datasets or performing highly scalable deductive reasoning.

Our proposal paves the way for a tight integration between advancements in the Knowledge Representation field and advancements in the NeSy field through the use of circuits as a unifying framework. 

As described in Section \ref{sec:dl-circuit}, the attractive properties of circuits can be used to implement methods beyond NeSy approaches, tackling real-world problems where domain knowledge can be formalized as an ontology. In this work, we preliminarily explored some applications (data generation and deductive reasoning) to showcase the advantages of compiling a circuit $\Co$. Future works include experimenting with $\Co$ on real-world use cases, using datasets and benchmarks that closely resemble real environments.

Nonetheless, this approach entails facing the main limitations of our proposal: the scalability of the compilation algorithm with respect to the size of the ontology. Future works include identifying other methods to encode DL fragments to more succinct circuit representations and exploring the limits of knowledge compilers with respect to the dimension of an ontology.

\begin{ack}
NL is supported by the FAIR – Future Artificial Intelligence Research Foundation as part of the grant agreement MUR n. 341, code PE00000013 CUP 53C22003630006.
AV is supported by the ``UNREAL: Unified Reasoning Layer for Trustworthy ML'' project (EP/Y023838/1) selected by the ERC and funded by UKRI EPSRC.
\end{ack}

\bibliography{bibliography}

\end{document}